\documentclass[conference]{IEEEtran}

\usepackage[numbers]{natbib}
\usepackage{multicol}
\usepackage[bookmarks=true]{hyperref}
\usepackage{amsmath}
\usepackage{amssymb}
\usepackage{color}

\usepackage{graphicx}
\graphicspath{{images/}}

\usepackage[dvipsnames]{xcolor}

\newcommand{\paral}{{\!/\mkern-5mu/\!}}

\newcommand{\ma}{\mathbf{a}}

\newcommand{\q}{\mathbf{q}}
\newcommand{\qd}{{\dot{\q}}}
\newcommand{\qdd}{{\ddot{\q}}}
\newcommand{\uu}{\mathbf{u}}
\newcommand{\ud}{{\dot{\uu}}}
\newcommand{\udd}{{\ddot{\uu}}}
\newcommand{\vv}{\mathbf{v}}

\newcommand{\x}{\mathbf{x}}
\newcommand{\xd}{{\dot{\x}}}
\newcommand{\xdd}{{\ddot{\x}}}
\newcommand{\y}{\mathbf{y}}

\newcommand{\z}{\mathbf{z}}
\newcommand{\zd}{{\dot{\z}}}

\newcommand{\mr}{\mathbf{r}}
\newcommand{\s}{\mathbf{s}}

\newcommand{\oo}{\mathbf{o}}
\newcommand{\f}{\mathbf{f}}
\newcommand{\h}{\mathbf{h}}
\newcommand{\g}{\mathbf{g}}

\newcommand{\mt}{\mathbf{t}}

\newcommand{\zero}{\mathbf{0}}

\newcommand{\Real}{\mathbb{R}}
\newcommand{\J}{\mathbf{J}}
\newcommand{\Jd}{{\dot{\J}}}
\newcommand{\Jp}{\J_{\phi}}
\newcommand{\Jdp}{\Jd_{\phi}}
\newcommand{\A}{\mathbf{A}}
\newcommand{\B}{\mathbf{B}}

\newcommand{\D}{\mathbf{D}}

\newcommand{\G}{\mathbf{G}}
\newcommand{\mH}{\mathbf{H}}
\newcommand{\I}{\mathbf{I}}
\newcommand{\mL}{\mathbf{L}}
\newcommand{\M}{\mathbf{M}}
\newcommand{\U}{\mathbf{U}}
\newcommand{\mR}{\mathbf{R}}
\newcommand{\mS}{\mathbf{S}}
\newcommand{\mT}{\mathbf{T}}
\newcommand{\V}{\mathbf{V}}
\newcommand{\W}{\mathbf{W}}

\newcommand{\inner}[2]{\langle#1,#2\rangle}
\newcommand{\wt}[1]{\widetilde{#1}}

\newcommand{\T}{\top}

\newcommand{\calQ}{{\cal Q}}
\newcommand{\calC}{{\cal C}}
\newcommand{\calR}{{\cal R}}
\newcommand{\calU}{{\cal U}}
\newcommand{\calX}{{\cal X}}

\usepackage{amsmath}  
\usepackage{amssymb}
\usepackage{accents}  
\usepackage{amsthm}

\newcommand{\ubar}[1]{\underaccent{\bar}{#1}}

\newcommand{\myeqref}[1]{Eq.~\eqref{#1}}

\theoremstyle{plain}
\newtheorem{lemma}{Lemma}[section]

\newtheorem{corollary}{Corollary}[section]

\theoremstyle{definition}

\theoremstyle{remark}

\let\labelindent\relax  

\usepackage{amsmath}
\usepackage{amsfonts}
\usepackage{amssymb}
\usepackage{amsthm}
\usepackage{bm}
\usepackage{bbm}
\usepackage{mathtools}
\usepackage{enumitem}
\usepackage{thmtools,thm-restate}
\usepackage{algorithm}
\usepackage{algorithmic}
\usepackage{color}
\usepackage{graphicx}
\usepackage{comment}

\def\Rbb{\mathbb{R}}

\def\R{\Rbb}

\def\*{\star}

\DeclareMathOperator*{\argmin}{arg\,min}

\begin{document}

\title{Riemannian Motion Policies}
\date{}


\author{
\authorblockN{Nathan D. Ratliff}
\authorblockA{NVIDIA \\
{\small nratliff@nvidia.com}}
\and
\authorblockN{Jan Issac}
\authorblockA{NVIDIA \\
{\small jissac@nvidia.com}}
\and
\authorblockN{Daniel Kappler}
\authorblockA{Max Planck for Intelligent Systems \\
{\small daniel.kappler@tue.mpg.de}}
\and
\authorblockN{Stan Birchfield}
\authorblockA{NVIDIA \\
{\small sbirchfield@nvidia.com}}
\and
\authorblockN{Dieter Fox}
\authorblockA{NVIDIA \\
{\small dieterf@nvidia.com}}
}



%

\maketitle

\begin{abstract}
We introduce the Riemannian Motion Policy (RMP), a new mathematical object for modular motion generation.  An RMP is a second-order dynamical system (acceleration field or \textit{motion policy}) coupled with a corresponding Riemannian metric.  The motion policy maps positions and velocities to accelerations, while the metric captures the directions in the space important to the policy. We show that RMPs provide a straightforward and convenient method for combining multiple motion policies and transforming such policies from one space (such as the task space) to another (such as the configuration space) in geometrically consistent ways. The operators we derive for these combinations and transformations are provably optimal, have linearity properties making them agnostic to the order of application, and are strongly analogous to the covariant transformations of natural gradients popular in the machine learning literature. The RMP framework enables the fusion of motion policies from different motion generation paradigms, such as dynamical systems, dynamic movement primitives (DMPs), optimal control, operational space control, nonlinear reactive controllers, motion optimization, and model predictive control (MPC), thus unifying these disparate techniques from the literature. RMPs are easy to implement and manipulate, facilitate controller design, simplify handling of joint limits, and clarify a number of open questions regarding the proper fusion of motion generation methods (such as incorporating local reactive policies into long-horizon optimizers).  We demonstrate the effectiveness of RMPs on both simulation and real robots, including their ability to naturally and efficiently solve complicated collision avoidance problems previously handled by more complex planners.

\end{abstract}


\section{Introduction}

Robot motion generation today is extremely complex with many moving parts. On one side of the spectrum we have {\it collision-complete} motion planning \cite{LavallePlanningAlgorithms06}, tasked with finding a connected global collision-free path from start to goal, and at the other end we have purely reactive motion that reacts locally and quickly \cite{parkPastor2008AdaptingDMPs,haddadin2010iros,flacco2012icra,Khansari_Billard_AR_12,kaldestad2014icra}. Between those two extremes we have motion optimization \cite{RatliffCHOMP2009,ToussaintTrajOptICML2009,ParkITOMP2012,AbbeelTrajopt2013,KOMOToussaint2014,RIEMORatliff2015ICRA,Mukadam-ICRA-17}, which leverages powerful optimizers to smooth out plans, push them away from obstacles, implement local anticipatory behaviors for improved efficiency and naturalness, and better coordinate different goals. But this complexity is unsatisfying. Clearly, there’s a gap between the computational complexity of these tools and the seeming simplicity with which humans solve many of these tasks. Humans are innately reactive in their movements, solving and adapting these motion problems extremely fast.

In this work, we show that many seemingly complex motion generation problems, ones that seem easy and commonplace for humans but which are highly constrained and therefore difficult for modern planners and optimizers, are actually much simpler than previously believed and often solvable by purely local reactive motion policies. Many researchers intuited between the mid-2000’s and early 2010’s (see \ref{sec:relatedWork}) that an effective obstacle avoidance technique was to simply reach toward the target and locally perturb the arm away from nearby obstacles. But often in practice those techniques were less effective and less practical than expected--they would be unstable and oscillate at times, they would slow down near obstacles, and generally the large collection of obstacle policies and task policies would fight with each other and require special nullspace engineering to work well with one another. This paper revisits these local techniques by returning to the question of how to most effectively combine many, often competing, local motion policies into a single motion generation policy. We show that to optimally combine multiple policies we need to be careful about tracking how their local geometry (the directions of importance to the policy) transforms between spaces (such as when transformed from the end-effector space back to the configuration space) and how those local geometries contribute to their combination into a single policy.

\begin{figure}[t]
\begin{center}
\includegraphics[width=.85\columnwidth]{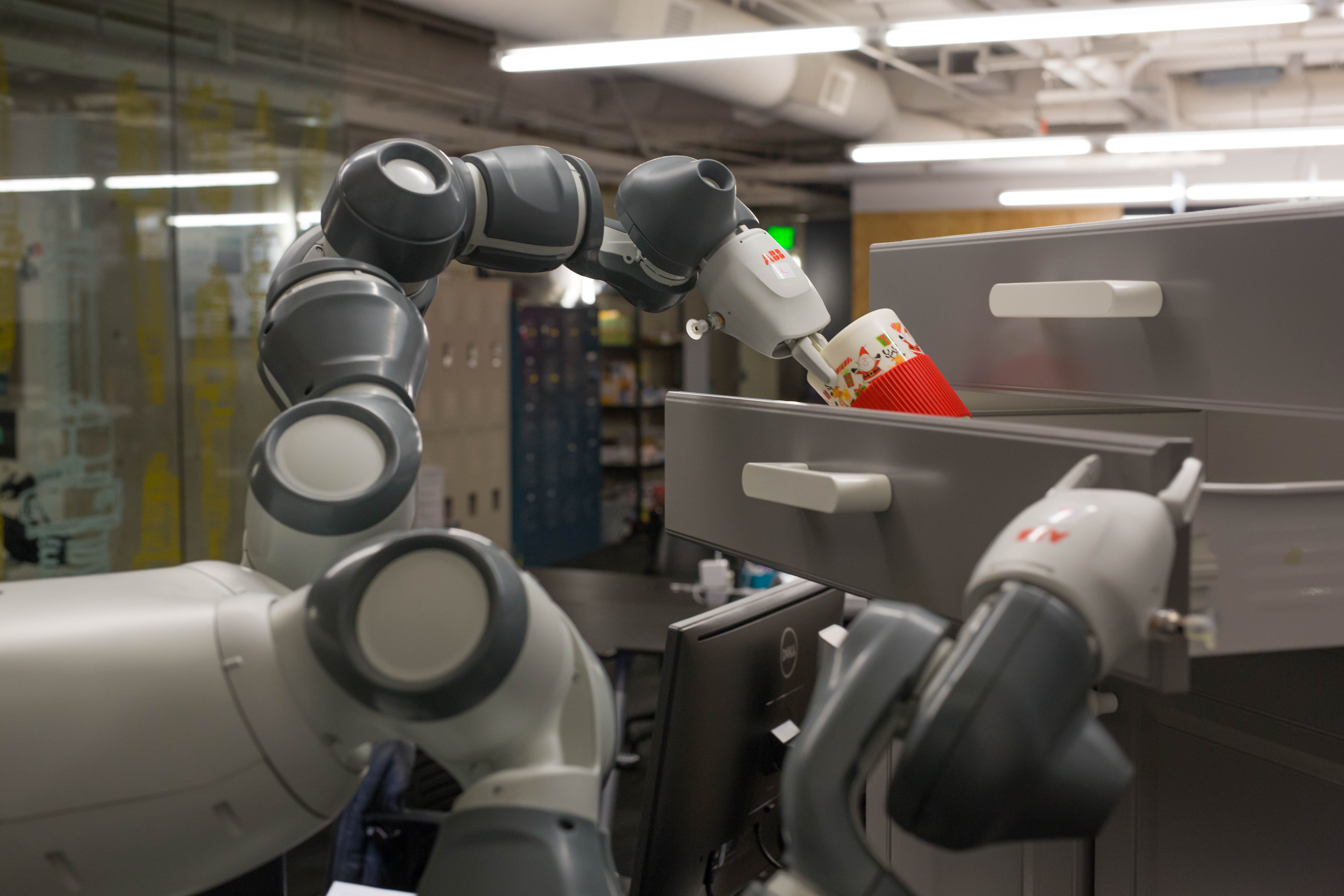}
\end{center}
\caption{A motivating application for this work is operating collaboratively in human environments, which are unstructured and highly dynamic. Here a two-armed ABB YuMi robot opens a drawer and inserts a cup, while avoiding unintended collisions.  To solve such problems, fast reactive local control is  critical. In this work, we present a framework for incorporating agile, reactive, highly adaptive behaviors in a modular way. The resulting motion generation system aggregates hundreds of controllers, is straightforward to implement, and performs well in practice.}
\label{fig:yumiexample}
\end{figure}

To formalize these considerations, we define a new mathematical object called the Riemannian Motion Policy (RMP) by pairing each policy with a Riemannian metric that defines its local geometry as a smoothly changing symmetric positive (semi-)definite matrix. We derive operators for the geometrically consistent transformation and combination of RMPs and prove their associativity and optimality. These operators behave analogously to natural gradient operations from the machine learning literature \cite{AmariNaturalGradients1998,BagnellCovariantPolicySearch2003}. More, working with them is as straightforward and modular as working directly with dynamical system policy representations with pseudoinverses and superposition, two common tools from the literature. But rather than using pseudoinverses we define geometrically consistent pullback operations, and rather than simply adding together the policies we perform metric-weighted averages of multiple policies which account for each policy's local geometry. Getting these critical geometric operations right has a significant impact on the performance and behavioral consistency of the overall system.

We demonstrate our framework using three dual-arm manipulation platforms both in simulation and reality. We show that 1. local motion policies alone can solve surprisingly complex collision avoidance problems; 2. they can be used to simplify long-range navigation and robustly solve a large practical class of those problems; 3. we can merge MPC-style continuous motion optimization with these local controllers to automatically achieve more anticipatory and coordinated behavior; 4. such large-scale optimizers or related computationally intensive behavior generation methods can be run in a separate process and communicated as a stream of RMPs without loss of fidelity to design efficient motion systems with diverse computational requirements. 

We note that RMPs are designed to illuminate how many techniques from the literature can be effectively combined with one another. In that sense, we see the framework as encompassing the umbrella term of {\it motion generation} rather than either motion planning or reactive motion alone. Motion planners, whether Probabilistic Road Maps \cite{LavallePlanningAlgorithms06}, Rapidly-exploring Randomized Trees (RRTs) \cite{LavallePlanningAlgorithms06}, Bit* \cite{bit_star_Gammell-2015}, or others, can be used to seed (or re-seed) any of a number of different types of continuously running motion optimizers \cite{RatliffCHOMP2009,ToussaintTrajOptICML2009,ParkITOMP2012,AbbeelTrajopt2013,KOMOToussaint2014,RIEMORatliff2015ICRA,Mukadam-ICRA-17}, and those results can be streamed as RMPs to the local reactive motion controllers which may, themselves, be Dynamic Movement Primitives \cite{IjspeertDMPs2013,parkPastor2008AdaptingDMPs}, Dynamical Systems \cite{Khansari_Billard_AR_12}, or custom collision controllers \cite{flacco2012icra,haddadin2010iros,kaldestad2014icra}. Despite the specifics of the motion techniques employed for each of the often multiple time-scales of a robotic system, RMPs provide a geometrically consistent framework for combining them all into a single motion policy within the configuration space, and show that the Riemannian metric is a critical additional quantity, alongside the policy itself, that must be tracked through transformations and combinations to ensure optimality and consistency of the result.

The Appendix provides a complete mathematical overview of RMPs, detailing many of 
specific RMPs used in the full system experiments.

\section{Related work} \label{sec:relatedWork}

Operational space control \cite{KhatibOperationalSpaceControl1987}, especially in its generalized form involving multiple task spaces as described in \cite{10-toussaint-BayesianControl}, forms a basis for the mathematics behind RMPs. That body of work, though, is largely independent of the literature around dynamical system representations such as the motion representations of Billard \cite{Khansari_Billard_AR_12} or Dynamic Movement Primitives (DMPs) \cite{IjspeertDMPs2013}. Those latter frameworks represent motion as second-order differential equations as we do here, but they do not address concretely how to combine multiple policies defined across many task spaces. 

Operational space control, on the other hand, addresses how to combine multiple  policies, but it does not address the problem of shaping the behavior of the policies, largely viewing the problem as that of instantaneous calculations. This leads to subtle but important ambiguities, such as how to define effective reactive collision controllers across the robot that account for many obstacles simultaneously in the space. As we show below, each collision policy at a given body point, itself, is a combination of many policies, one for each obstacle, making the metric at that body point important. Existing approaches often include a weight matrix on each task space term but usually do not justify the metric; in practice it is common to use naive metrics tailored to just one or two obstacles or tuned to perform only within specific environments. 

This work merges these two paradigms. For instance, it is common for dynamical systems to be defined at the end-effector and transformed to the configuration space using the pseudoinverse \cite{parkPastor2008AdaptingDMPs} where they are superimposed with one another. The reliance on the pseudoinverse is due to its utility as a computational tool to transform dynamical systems from one space to another. As we show here, the RMP framework allows us to define similar tools with similar utility in their modularity, innately behaving as in optimal operational space control. 

In the literature, there have been a number of frameworks proposed with similar computational results. For instance, operational space control defines the basic optimization-based formulation and the mathematics of representing each instance as a quadratic optimization. Similarly, \cite{10-toussaint-BayesianControl} and \cite{paraschos_prob_prio2017} augment these ideas to show that the mathematics of Gaussian inference plays a similar role; however, in many cases real probabilistic models are not available, so in practice they are often constructed manually to shape their behavior. Here we define the metric specifically as a tool for stretching the space in order to shape how different RMPs
interact with one another.  These metrics are globally defined across the entire state space and must be smoothly varying.

Importantly, none of these previous approaches discusses the recursive nature of the mathematics. By explicitly demonstrating how to calculate operational space control results in parts, the RMP framework decomposes the computation as needed across multiple computational devices or processes. For instance, in Section~\ref{sec:discussion} we explain how to keep the main high-speed RMP core computationally clean and reactive for handling fast integration of obstacle avoidance and reactive behaviors, separating it from the computationally demanding processes for intelligent behavior generation that interfaces to other areas of the perceptual and task processing system and may perform more sophisticated optimizations across future time horizons. The theory of RMPs ensures that all such components can be computed separately and combined into a single C-space linear RMP that captures all relevant information from these more complex behavioral components to reduce whatever needs to be communicated to the central RMP core. The central RMP core, similarly, only needs a single interface to receive those linear RMPs and combine them with the rest of the components being processed in the faster loop, thus enabling the careful control of computational requirements.

More broadly, the fields of optimal control \cite{OptimalControlEstimationStengel94} and model predictive control \cite{DRCIntegratedSystemTodorov2013} implicitly formalize these ideas of component combination by defining all components as nonlinear objective functions on various task spaces. However, in many cases, as above, behavior can be naturally described directly by differential equations (dynamical systems) without reference to an objective. More importantly, the framework of optimization alone does not provide us with useful tools for decomposing the problem across task spaces for behavioral design and reuse, or for distributing the computation across multiple processes or devices. For instance, as described above, optimal control can be computationally intensive and should be separated in practice from local reactive control as is done in \cite{2017_rss_system}. RMPs provides a natural framework and toolset for decomposing the problem such that the final result is optimal when combined.

\section{Mathematics of Robot Control}

We start by defining some basic concepts and notation.
Let $\q(t) \in \calQ \subset \Real^d$ denote the $d$-dimensional configuration of the robot at time $t$, i.e., the generalized coordinates of the system in the configuration space $\calQ$.  Typically $\q$ contains the joint angles, so that $\qd, \qdd$ are the velocities and accelerations of the joints, respectively. Similarly, let us assume that there is a set of non-linear task spaces, and let $\x_i(t) \in \calX_i \subseteq \Real^{k_i}$ denote the $k_i$-dimensional task variable in the $i^{\text{th}}$ task space $\calX_i$ at time $t$, with associated velocities and accelerations given by $\xd_i$ and $\xdd_i$.  The differentiable task map $\phi_i:\Real^d\rightarrow\Real^{k_i}$ relates the configuration space to the $i^{\text{th}}$ task space, so that $\x_i = \phi_i(\q)$.  For example, if $\x_i$ is the position and orientation of the end effector, then $\phi_i$ is the forward kinematics of the robot.  In the following, we drop the subscript for simplicity, and we often drop the explicit dependence upon time when the context is clear.

If we denote the Jacobian of a task map $\phi$ as
\begin{align}
	\Jp \equiv \frac{\partial\phi}{\partial\q} \in \Real^{k \times d},
\end{align}
then the task space velocities and accelerations are given by
\begin{align}
	\xd &= \frac{d}{dt}\phi(\q) = \Jp\qd \label{eq:xdeqjqd} \\
    \xdd &= \frac{d^2}{dt^2}\phi(\q) = \Jp\qdd + \Jdp\qd \approx \Jp\qdd, \label{eq:xddeqjqdd} 
\end{align}
where the last approximation drops the term associated with the second-order curvature of $\phi$,\footnote{Specifically, $\Jdp\qd$ is the curvature of $\phi$ in the direction of the velocity $\qd$.} similar to Gauss-Newton approximation. In practice, this second-order correction term is unnecessary because integration steps in control loops (running between 100~Hz and 1~kHz) are small.



A Riemannian metric $\A$ is a symmetric positive semidefinite matrix defined on the tangent space that measures the inner product between two tangent vectors $\uu$ and $\vv$ as $\inner{\uu}{\vv}_\A = \uu^\T\A\vv$. Thus, $\|\uu\|_\A^2 = \uu^\T\A\uu$, which reduces to the squared Euclidean norm when $\A$ is identity.  

For a given Riemannian metric $\A$, from \myeqref{eq:xdeqjqd}--\eqref{eq:xddeqjqdd} we have $\xd^\T\A\xd = \qd(\J^\T\A\J)\qd$, and $\xdd^\T\A\xdd = \qdd(\J^\T\A\J)\qdd$, where we drop the subscript on the Jacobian when it is clear from the context.  Therefore,
\begin{align}
    \|\xd\|_\A^2 &= \|\qd\|_{\B}^2 \label{xdaqdb} \\
    \|\xdd\|_\A^2 &= \|\qdd\|_{\B}^2, \label{xddaqddb}
\end{align}
where $\B \equiv \J^\T\A\J$ is the metric in the domain of the map that mimics the metric $\A$ in the codomain.  As we shall see, the ease with which a metric can be transformed between the domain and codomain is key to the unique properties of the proposed approach.

When the state space consists of position and velocity, it is $2k$-dimensional; nevertheless, only a $k$-dimensional slice of the tangent space is consistent with the integration equations, 
$\x({t+1}) = \x(t) + \xd(t) \, \Delta t$, and $\xd({t+1}) = \xd(t) + \xdd(t) \, \Delta t$, 
where $\Delta t$ is the timestep.  As a result, we restrict the following to the $k$-dimensional subspace of the tangent space identifiable with the acceleration vector $\xdd$, and therefore we consider only Riemannian metrics $\A(\x, \xd)\in \Real^{k\times k}$ defined on this subspace. 

We define a \emph{motion policy} $\f : \q, \qd \mapsto \qdd$ as a dynamical system (second-order differential equation) mapping position and velocity to acceleration. 
However, since the configuration space is equivalent to a task space if $\phi$ is the identity map, we can write $\f : \x, \xd \mapsto \xdd$ without loss of generality.  As we shall see, the latter is often more convenient, since it is more natural to define all motion policies in the task space.  

Unlike search- or optimization-based motion planning approaches that represent the control by a single trajectory  \cite{LavallePlanningAlgorithms06,Kalakrishnan_RAIIC_2011,AbbeelTrajopt2013}, a motion policy encodes an infinite bundle (continuum) of trajectories in its integral curves in the sense that for any initial state, the policy generates a trajectory through forward integration. Using Euler integration, for example, the trajectory is generated as
\begin{align}
    \q({t+1}) &= \q(t) +  \qd(t) \, \Delta t \\
    \qd({t+1}) &= \qd(t) + \f(\q(t), \qd(t)) \, \Delta t.
\end{align}
The robot's underlying control system might consume any subset of the resulting triplet $(\q, \qd, \qdd)$ of position, velocity, and acceleration at each time step, depending on whether it is position controlled, velocity controlled, position and velocity controlled with inverse dynamics, and so forth. 

In reality, control systems are often capable of controlling the position of the robot with high precision using high-gain PID controllers. Therefore, reactions and compliance can often be implemented directly as modifications to the position signals.  However, when contact with the environment is important, force and/or torque control can be integrated.  The framework presented in this paper is sufficiently general to handle all these cases.

\section{Riemannian Motion Policies (RMPs)}

In this section we introduce a new approach to representing and transforming motion policies.  We show that this approach allows motion policies to be combined in a way that preserves their geometry, thus leading to a provably optimal control system that readily transfers from one robot to another without re-tuning of parameters.

We define a \emph{Riemannian Motion Policy (RMP)}, denoted by the tuple ${^\calX}(\f, \A)$, as a motion policy (or dynamical system) in a space $\calX$ augmented with a Riemannian metric. The dynamical system $\f : \x, \xd \mapsto \xdd^d$, where $\x \in \calX$, is described by a second-order differential equation mapping position and velocity to a desired acceleration $\xdd^d = \f(\x,\xd)$.  For this reason, it is also known as an \textit{acceleration policy}. Typically, both the differential equation and task space are nonlinear.  The Riemannian metric $\A(\x, \xd)$ is a positive semidefinite matrix that varies smoothly with the state $(\x, \xd)$.

\subsection{Operators}

We now describe several important operators of RMPs, along with their properties; derivations can be found in the appendix. These operators define how RMPs transform between the different spaces of a robotic system (e.g. between the end-effector space and the configuration space) and how multiple RMPs should combine with one another in a way that trades off their respective geometries.

\textbf{Addition.}  
If $\calR_1={^\Omega}(\f_1,\A_1)$ and $\calR_2={^\Omega}(\f_2,\A_2)$ are two RMPs in some space $\Omega$, then they naturally combine in the following manner:
\begin{align} \label{eqn:RMPCombinationTwo}
    \calR_1+\calR_2 = \left((\A_1+\A_2)^{+}(\A_1\f_1+\A_2\f_2),\ \A_1+\A_2\right)_\Omega,
\end{align}
where $^+$ is the pseudoinverse, which reduces to the inverse when the matrix is full rank. More generally, a collection of RMPs $\{\calR_i\}_{i=1}^n$ are combined into a single RMP $\calR_c={^\Omega}(\f_c,\A_c)$ as a \textit{metric-weighted} average:
\begin{align} \label{eqn:RMPCombination}
    \calR_c = \sum_i \calR_i = {^\Omega}\left(\left(\sum_i\A_i\right)^{+}\sum_i\A_i\f_i,\,\, \sum_i\A_i\right)
\end{align}
that yields the optimal solution to the combined system.
Note that if each metric is of the form $\A_i = w_i\I$ (i.e., the eigenspectrum is axis-aligned), this reduces to the traditional weighted average: $(\f_c, \A_c) = \big(\frac{1}{w}\sum_iw_i\f_i, \frac{1}{w}\sum_iw_i\A_i\big)$, where $w \equiv \sum_iw_i$.

\textbf{Pullback.} In differential geometry, the \emph{pullback} of a function $\f$ by a mapping $\phi$ satisfies $\mathtt{pull}_\phi(\q) = \f(\phi(\q))$.  In other words, the pullback applied to the domain yields the same result as the function applied to the co-domain.  Similarly, the pullback operation of an RMP ${^\calX}(\f,\A)$ defined in the co-domain $\calX$ is an equivalent RMP defined in the domain $\calQ$:
\begin{align} \label{eqn:pullback}
    \mathtt{pull}_\phi\left({^\calX}(\f, \A)\right)= {^\calQ}\left(\left(\J^\T\A\J\right)^{+}\J^\T\A\f,\ \J^\T\A\J\right),
\end{align}
where the transformation of $\A$ into $\J^\T\A\J$ is the same as we saw in \myeqref{xddaqddb}.  If $\J$ is full row rank, then the expression for the transformation of $\f$ simplifies to 
\begin{align} \label{eqn:pullbacksimple} \mathtt{pull}_\phi\left({^\calX}(\f, \A)\right)= {^\calQ}\left(\J^+\f,\ \J^\T\A\J\right).
\end{align}

There is a strong connection between \myeqref{eqn:pullback} and the natural gradient common in machine learning \cite{AmariNaturalGradients1998,BagnellCovariantPolicySearch2003}. If $\nabla_{\xdd}\mathcal{F}(\xdd; \x, \xd) = \f(\x, \xd)$ is the gradient of a potential function $\mathcal{F}(\xdd; \x, \xd)$, the parametric gradient of the composed function $\psi:\q \mapsto \x$ is $\nabla_{\qdd} \psi(\J^\T\xdd; \x, \xd) = \J^\T\A\f$. The differential equation transformation defined by \myeqref{eqn:pullback} is, therefore, the parametric gradient transformed by the pullback metric $\J^\T\A\J$. Thus, we can view this pullback differential equation as the \textit{natural vector field}, in analogy to the natural gradient.

\textbf{Pushforward.}  Similarly, we can transform an RMP from the domain of a task map to its co-domain.  The \emph{pushforward} operation applied to an RMP $(\h,\B)$ defined on the domain $\calQ$ yields an equivalent RMP defined on the co-domain $\calX$:
\begin{align}
    \mathtt{push}_\phi\left({^\calQ}(\h, \B)\right) = {^\calX}\left(\J\h, \,\,{(\J^{+})}^\T\B\J^+\right),
\end{align}
where $\B=\J^\T\A\J$ as before.  As an example, if $\h$ is defined on the configuration space $\calQ$, and if $\phi$ is a forward kinematics map to the end-effector space $\x = \phi(\q)$, then the pushforward of $\h$ by $\phi$ describes the end-effector movement under $\h$ along with its associated metric.

\subsection{Properties}

\textbf{Commutativity and associativity of addition.}  Addition of RMPs is both commutative and associative.  That is, 
\begin{align}
	\calR_1 + \calR_2 &= \calR_2 + \calR_1 \\
    (\calR_1 + \calR_2) + \calR_3 &= \calR_1 + (\calR_2 + \calR_3)  
\end{align}
for any RMPs $\calR_1$, $\calR_2$, and $\calR_3$ in the same space. 

\textbf{Linearity of pullback and pushforward.}
The pullback and pushforward operators are both linear:
\begin{align}
 \mathtt{pull}_\phi(\calR_1+\calR_2) &= \mathtt{pull}_{\phi}(\calR_1)+\mathtt{pull}_{\phi}(\calR_2) \\
 \mathtt{push}_\phi(\calR_1+\calR_2) &= \mathtt{push}_{\phi}(\calR_1)+\mathtt{push}_{\phi}(\calR_2).
\end{align}

\textbf{Associativity of pullback and pushforward.}
The operators are both associative.  Let $\z = \phi_1(\q)$ and $\x = \phi_2(\z)$ so that $\x = (\phi_2\circ\phi_1)(\q) = \phi_2(\phi_1(\q))$ is well defined, and suppose $\calR$ is an RMP on $\calX$ and $\calR'$ is an RMP on $\calQ$. Then
\begin{align}
 \mathtt{pull}_{\phi_1}\big(\mathtt{pull}_{\phi_2}(\calR)\big) &= \mathtt{pull}_{\phi_2\circ\phi_1}(\calR) \\
 \mathtt{push}_{\phi_2}\big(\mathtt{push}_{\phi_1}(\calR')\big) &= \mathtt{push}_{\phi_2\circ\phi_1}(\calR').
\end{align}

\textbf{Covariance of pullback and pushforward.}
Like related operations in Riemannian geometry, the pullback operation is covariant to reparameterization \cite{AmariInformationGeometry1994,BagnellCovariantPolicySearch2003}, which means that it is unaffected by a change of coordinates. Specifically, let $\q = \zeta(\uu)$ be a bijective differentiable map with Jacobian $\J_\zeta$, and let $\calR={^\calQ}(\f, \A)$ be an RMP on $\calQ$.  Then 
\begin{align}
    \mathtt{pull}_\zeta\left({^\calQ}(\f, \A)\right) = {^\calU}\left(\J_{\zeta}^\T(\J_{\zeta}\J_{\zeta}^\T)^{+}\f,\ \J_\zeta^\T\A\J_\zeta\right) \label{eq:covarianttransform}
\end{align}
is equivalent to $\calR$ up to numerical precision. That is, for any $(\uu, \ud)$, the following holds:
\begin{align}
    \J_\zeta^\T\h(\uu, \ud) = \f(\q, \qd),
\end{align}
where $\q=\zeta(\uu)$, $\qd=\J_\zeta\ud$, and $\h:(\uu,\ud) \mapsto \udd$ is given by the transformed differential equation in \myeqref{eq:covarianttransform}, i.e., the first element in the tuple.  The importance of this result is that  integral curves created in $\calU$ and transformed to $\calQ$ will match the corresponding integral curve found directly in $\calQ$. (The same holds true for pushforward.)

This covariance property makes it extremely convenient to handle joint limits as described in Section~\ref{sec:JointLimits}. At a high-level we can pull the constrained C-space RMP back through an inverse sigmoid into an unconstrained space that always satisfies the joint limits.


\section{Motion Generation}

RMPs can be used to generate the motion of the robot in a straightforward manner, elegantly combining the various contributions into a single dynamical system.  The robot is typically modeled by a collection of points $\x_1,\x_2,\ldots$ on the robot's body, with associated forward kinematics functions $\x_i = \phi_i(\q)$.  In the following we assume that, for any interesting behavior, the robot will be governed by a collection of RMPs $\{\calR_1, \calR_2, \ldots \}$ with associated dynamical systems and Riemannian metrics.  (This notation is not meant to imply a one-to-one correspondence between RMPs and robot points.)  
In this section we describe how multiple RMPs in various task spaces can be combined to find an optimal RMP that can be executed on the robot, followed by several basic local reactive policies, and finally an analysis of some of the properties of this approach.

\subsection{Combining RMPs to solve motion generation}

Suppose we have a set of task maps $\phi_i$ with associated desired acceleration vector fields $\xdd_i^d$ with Riemannian metrics $\A_i$.  Our goal is to find a motion policy in configuration space $\f:\q,\qd \mapsto \qdd$ such that
\begin{align}
    \f(\q,\qd) = \arg \min_{\qdd} \sum_i \frac{1}{2}\|\xdd_i^d - \underbrace{\J_{\phi_i}\qdd}_{\xdd_i}\|_{\A_i}^2. \label{eq:quadraticsum}
\end{align}
That is, we wish to find the second-order dynamical system that minimizes the cost function combining all the desired accelerations while taking into account their associated metrics.

The RMP framework solves this problem in a straightforward manner via the following steps:  1) an RMP ${^{\calX_i}}(\f_i,\A_i)$ is created for each task map, where $\f_i \equiv \xdd_i^d$; 2) the RMPs are pulled back into the configuration space using \myeqref{eqn:pullback}; 3) the pulled-back RMPs are summed using \myeqref{eqn:RMPCombinationTwo}; and 4) the combined RMP is itself pulled back into an unconstrained space to handle joint limits, which is explained in more detail at the end of this section.

 
From \myeqref{eq:quadraticsum}, it is easy to see that each RMP $(\f_i,\A_i)$ is associated with its corresponding term in the equation, namely,
\begin{align}
\frac{1}{2}\|\xdd_i^d - \underbrace{\J_{\phi_i}\qdd}_{\xdd_i}\|_{\A_i}^2. \label{eq:quadraticsingle}
\end{align}
Such a quadratic is parameterized by its 
vector field $\f_i$ (the minimizer of the quadratic) and its metric $\A_i$ (the Hessian of the quadratic).  In other words, we can think of $(\f_i,\A_i)$ as a compact notation for the quadratic term in \myeqref{eq:quadraticsingle}, similar to the way the mean and variance can be used to parameterize a Gaussian.


\subsection{Basic local reactive policies}

To better understand how this approach works in practice, in this section we detail several possible RMPs.  These are just examples --- many more possibilities exist.

\textbf{Target.} Perhaps the simplest RMP is one that attempts to pull a point on the robot toward a goal.  Let $\x = \phi_e(\q)$ be the 3D end-effector position, and let $\x_g$ be a desired target. Then a target controller can be defined as 
\begin{align}
    \f_g(\x, \xd) = \alpha \s(\x_g - \x) - \beta \xd,
\end{align}
where $\alpha,\beta > 0$ are scalar gains, and $\s$ is a soft normalization function: $\s(\vv) = \frac{\vv}{h(\|\vv\|)}$, where
\begin{align}
    h(z) = z + c \log(1 + \exp(-2 c z)),\ \ c > 0,
\end{align}
so that $h(z) \approx z$ if $z \gg c$, but $h(z) \rightarrow c\log(2) > 0$ when $z\rightarrow 0$ to avoid dividing by zero.

This policy pulls the end effector directly toward the target with force proportional to the error, and with damping proportional to velocity.  This results in well-behaved convergent behavior at the target, where the force is diminished to zero and damped accordingly.  The corresponding metric can be either the identity, $\A_e=\I$, or it can be directionally stretched; in practice we have found that both work well.

\textbf{Orientation.}  
Orientation controllers are often implemented by applying SLERP metrics to quaternions, but such an approach has no way of expressing partial constraints.  Oftentimes, however, we wish to enforce orientation constraints only along one of the axes.  With RMPs, this is easy, by simply applying target controllers (explained above) on a canonical point along the appropriate axis (e.g., the endpoint of the unit vector starting from the origin).  Or, by applying target controllers along two axes, the full orientation constraint can be applied.

\textbf{Collision.}  
Let $\oo_1,\oo_2,\ldots$ be a collection of obstacle points (e.g., key points indicating the closest points). For each $(\x_i, \oo_j)$ pair, we define a dynamical system and associated metric as
\begin{align}
    \f_{ij}(\x_i, \xd_i) &= 
        \alpha(d_{ij}) \widehat{\vv}
        - \beta(d_{ij}) \left(\widehat{\vv}\widehat{\vv}^\T\right) \xd_i \\
        \A_{ij} &= w(d_{ij})\s(\xdd_{ij})\s(\xdd_{ij}^\T)
\end{align}
where $\vv = \x_i - \oo_j$, $\widehat{\vv}=\vv/\|\vv\|$, $d_{ij}$ is the distance from $\x_i$ to $\oo_j$, and $\alpha(\cdot)$, $\beta(\cdot)$, and $w(\cdot)$ are scalar functions, with the weighting function $w$  decreasing to zero further from the obstacle.  Although we typically use the Euclidean distance ($d_{ij}= \|\vv\|$), the controllers are agnostic to the choice of distance function, so that alternative nonlinear distances, such as electric potentials  \cite{MainpriceWarpingIROS2016}, can also be used.

This policy both pushes the robot away from the obstacle and dampens the velocities in the direction of the obstacle.  Since the  controller does not care about motion orthogonal to $\vv$, the metric above is directionally stretched. Note that obstacles include world obstacles as well as the robot's own body (including its other arm). All obstacles are treated in the same way. Handling each obstacle as a separate RMP and using the RMP combination operations to combine the (many) different policies makes a substantial difference in practice.  


\textbf{Redundancy resolution and damping.} 
For redundant manipulators (i.e., those containing more than 6 degrees of freedom), we resolve the redundancy using a controller of the form
\begin{align}
    \f_d(\q, \qd) &= \alpha (\q_0 - \q) - \beta \qd\\
    \A_d &= \I,
\end{align}
where $\alpha,\beta > 0$ are constant gains and $\q_0$ is some default posture configuration. The identity metric is sufficient for this purpose. 

This controller is not covariant, because it is defined in the configuration space rather than the task space. As a result, it is robot-dependent. However, although this controller should ideally be defined using points on the arm, in practice it represents such a relatively minor contribution to the overall system behavior that we have found it acceptable to define it in the domain rather than the co-domain (thereby assuming $\phi$ is identity), which removes the dependence upon the robot.


These controllers are all local: they react without planning. They are very effective for local navigation among obstacles, but more sophisticated techniques are needed to guide the arm across long distances that require deciding how to navigate large obstacles that significantly warp the workspace geometry. 

\subsection{Integrating computationally intensive behaviors} \label{sec:CombiningWithMotionOptimization}

In addition to purely local behaviors such as those above, it is easy to define controllers that account for a short horizon anticipation within the RMP framework.  Leveraging ideas from optimal control and model predictive control (MPC) \cite{DRCIntegratedSystemTodorov2013}, we define a controller as the mapping from a state $(\q, \qd)$ to the first acceleration $\qdd$ along the locally optimal trajectory starting from the state, where optimality is defined in terms of some time-varying collection of cost functions.

Let $\qdd_{\mathrm{opt}} = \f_{\mathrm{opt}}(\q, \qd)$ denote the mapping from state to the optimized next action that would be taken by an MPC system. To properly account for the Riemannian geometry of the nonlinear manipulator and workspace, we leverage the Riemannian Motion Optimization (RieMO) framework \cite{RIEMORatliff2015ICRA}.  Under RieMO, the Gauss-Newton Hessian approximation, which is a pullback metric, defines a Riemannian metric $\A(\q, \qd)$ associated with the policy.  The pair yields an RMP,  ${^\calQ}(\f_{\mathrm{opt}},\ \A(\q, \qd))$, which can be combined with other RMPs.

In practice, evaluating $\f_{\mathrm{opt}}$ is expensive (requiring an optimization over a finite-horizon trajectory), and even with warm starts can only be evaluated at a low rates (e.g., 10-20 times a second) for reasonably sophisticated problems. Therefore, we use the typical optimal control technique of calculating linearizations of the problem, which can be re-evaluated at a much faster rate. This results in a time-varying sequence of linear RMPs that are communicated to the local controllers and combined using the RMP framework. Although these controllers are local policies, they anticipate the future, thus helping to coordinate target and orientation controllers with each other, generating more sophisticated and effective local obstacle avoidance behaviors, coordinating with the other arm, and so forth.

More broadly, any collection of linearized RMPs can be can be pulled back and combined in the configuration space as a single linear RMP by applying the above RMP rules to the linear policy coefficients. Doing so is convenient for funneling communication from external behavior generation processes down into a single RMP message to reduce communication and simplify integration with the rest of the reactive local controllers.

The supplementary video shows the performance of replacing the target controller described here with a continuous optimizer to boost coordination and anticipation.

\subsection{Heuristic long-range arm navigation} \label{sec:longRange}

Despite its simplicity, the RMP framework enables control over long ranges across the workspace. We describe here a simple but effective approach for long-range navigation that requires no planning as long as the robot's elbow is not blocked.\footnote{This assumption is reasonable in practice. Human work environments are often engineered to keep elbows free of obstacles.} Intuitively, this stems from the surprising effectiveness of simple retraction heuristics that pull the arm back to a retracted configuration from an outstretched pose (among the environment's obstacles---see Figure \ref{fig:experimentalEnvs}). We describe two such retraction heuristics here.

In describing this technique, we assume a 7-DOF robot with the common joint layout: an axis-aligned 3-DOF wrist, an axis-aligned 3-DOF shoulder, and a 1-DOF elbow.  Let $\q_r$ be some canonical, retracted configuration with the elbow back and the end-effector in a ready position near the robot's side. If the robot is in an outstretched position and needs to reach another outstretched position such that the obstacles between the two are insurmountable for the local controllers, all that is needed is for the robot to retract itself to the canonical configuration $\q_r$, then stretch out to reach the goal. 

In Section~\ref{sec:experimentsRetraction} we experiment with two retract heuristics. The first is a very simple attractor policy in the configuration space $\qdd = \alpha(\q_r - \q)$ pulling toward the retract configuration. The second is a slightly more sophisticated and robust retraction policy that additionally pulls the wrist toward a point the forearm (moving with the forearm as the retraction unfolds). Over the we eventually blends the target point toward where the wrist should be once retracted as a function of proximity to that point. The first heuristic is already quite robust---it is what we use in practice for many of our manipulation problems. But the second performs slightly better on some intentionally difficult environments studied in the experimental section.


For reaching forward, a simple heuristic is to calculate an inverse kinematics (IK) goal solution, apply one of the above retract heuristics, and play it backward to guide the arm across the obstacles. In practice, for increased flexibility, we apply a related heuristic whereby we follow a series of \textit{guiding points} that pull the arm into the general homotopy class it needs to reach the desired end-effector target. For this heuristic, we do not need full inverse kinematics solutions, just a way of generating effective guiding points. That said, in practice, we often do generate rough IK approximations which we call \textit{guiding configurations} and use points along the arm of the approximate IK solution as guiding points. We have found this simple heuristic to work extremely well for a wide range of practical problems. More global behaviors such as the effective choice of homotopy class (e.g., left vs.\ right around an obstacle) can be handled in a similar manner.

\subsection{Handling joint limits} \label{sec:JointLimits}



We can handle joint limits by defining a mapping between the constrained joint limit space and an unconstrained space using the sigmoid function. The intuition is that we pull the final combined joint limit constrained RMP into an unconstrained space by pulling it through a nonlinear sigmoidal map that maps the entire real line to the joint limit interval. This unconstrained space is convenient since simple affine policies (such as positional attractors) manifest as highly nonlinear policies in the original space after being transformed through the sigmoid. Because of space limitations, we defer the details to the appendix.





\section{Discussion}
\label{sec:discussion}
To appreciate the power of the RMP framework, it is helpful to contrast it with  alternatives.  Collision controllers, for example, are often described as motion policies, with contributions from multiple obstacles simply superimposed (added together) to form a joint controller \cite{IjspeertDMPs2013,parkPastor2008AdaptingDMPs}. But a simple thought experiment shows that superposition may result in unintuitive behavior.  Consider a scenario in which the robot reaches between two obstacles.
When the desired acceleration contributions from the two obstacles are symmetric then their sum is zero, in which the control system ignores the obstacles.  
On the other hand, if the obstacles are on the same side, then superposition inflates their contributions, while averaging them downweights important contributions. Such simple techniques all result in unintuitive artifacts and require per-scenario tuning for good performance. 

Another problem occurs when combining repelling forces to avoid obstacles with an attractive force to pull the end effector toward the goal.  The typical approach of superposition involves a weighted sum of these two competing desires.  To ensure obstacle avoidance, the weight on the obstacle terms is generally set to a high value, oftentimes resulting in sluggish behavior near obstacles as the robot slows down to avoid collision.  

By taking directionality into account, the RMP framework avoids all these problems.  Each obstacle point contributes both a force pushing the robot away from the point as well as a force slowing down the robot near the point, but only in the direction of the obstacle; that is, velocities in the plane perpendicular to the obstacle direction are unaffected, thus allowing the robot to glide smoothly around obstacles. 

Under the RMP framework, individual controllers can be  myopically designed to control only a small portion of the problem where the geometry is well understood. 
These individual controllers govern the distance from points on the robot to the closest point on an obstacle. 
Pullbacks through nonlinear maps connecting these spaces then transform the geometry to a common space where their contributions can be combined through metric-weighted averaging. 

Intuitively, the Riemannian metric defines (in its eigenspectrum) which directions in the space the policy cares about most.  The metric may thus  be viewed as a soft alternative to a null-space description, encoding trade-offs or preferences rather than hard directions of independence---although the framework is capable of encoding null-spaces if/when needed.

Because RMPs are geometrically consistent under transformation and combination (i.e., they are covariant to reparameterization), they are well-suited to parallel computation.  That is, different parts of the overall motion policy, such as separate long-horizon motion optimization and local reactive control components, can be computed separately to be combined later. 
This property also enables straightforward reuse of  policies across different robots. 

When parts of the system can only be run at a slower rate due to high computational complexity (e.g., motion planning), they can be computed offboard.  These nonlinear offboard motion policies can be linearized and communicated at a slower rate (e.g., 10~Hz) to a faster RMP core (e.g., a 1~kHz inner loop) to be integrated.  Since these policies are valid in a region around the linearization, the main RMP core can use them effectively during the time interval between policy updates.   This potential parallelization and separation of computation is another advantage of the RMP framework.

\section{Experiments}

In this section, we present some experimental results. We first demonstrate the degradation in performance of large systems of competing motion policies when the Riemannian geometry of the policies is not properly tracked during combination, and visualize some statistics of the eigenspectra of common pullback policies across trials of reaching tasks through clutter. We then show the success in long-range navigation of simple heuristics when using many strong local reactive controllers that are combined well within the RMP framework.

\subsection{Comparison to alternative techniques for C-space combination}

\begin{figure*}[t]
\begin{center}
\includegraphics[height=.30\columnwidth]{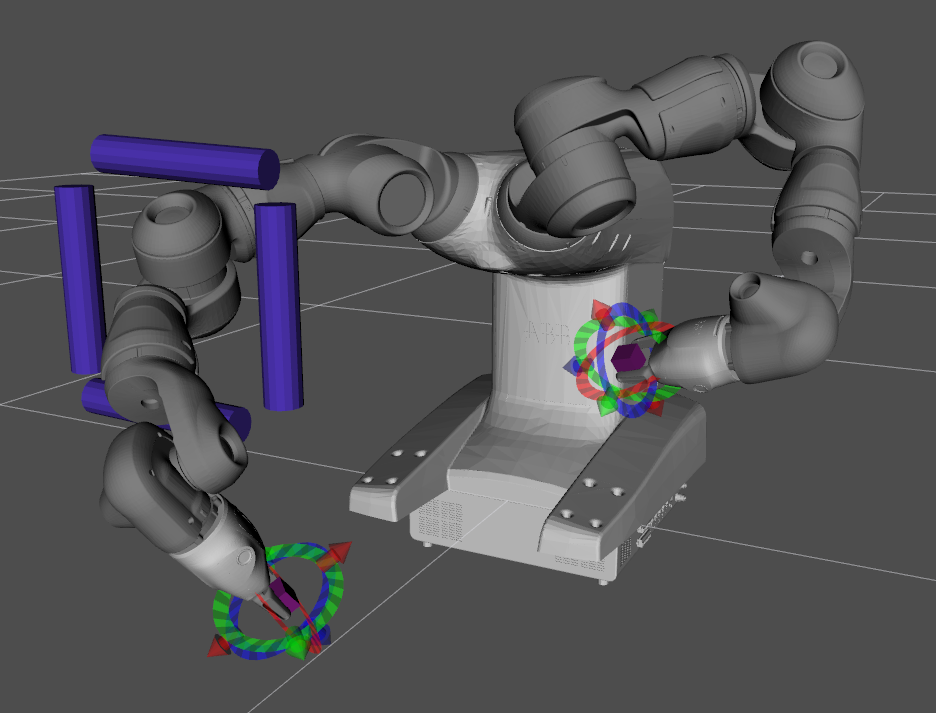}
\includegraphics[height=.30\columnwidth]{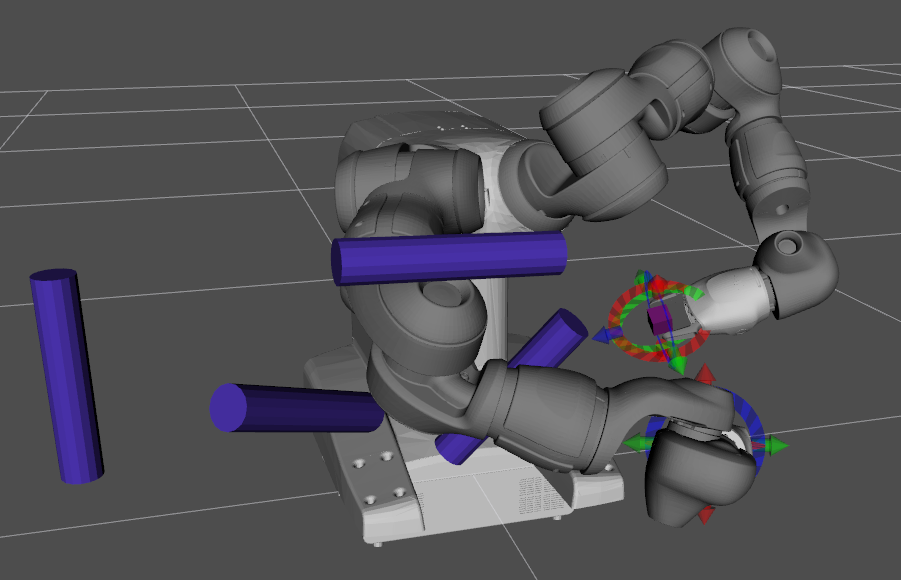}
\includegraphics[height=.30\columnwidth]{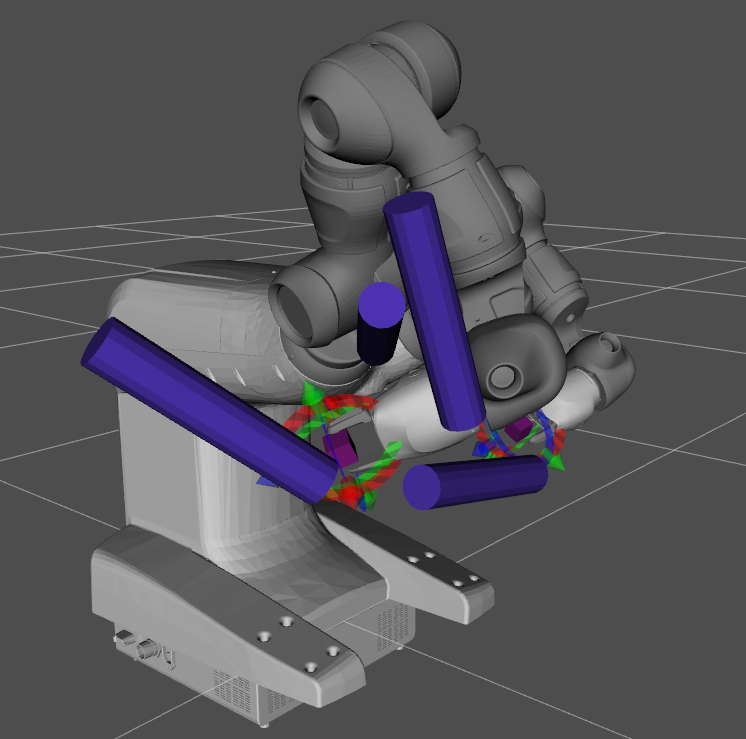}
\includegraphics[height=.30\columnwidth]{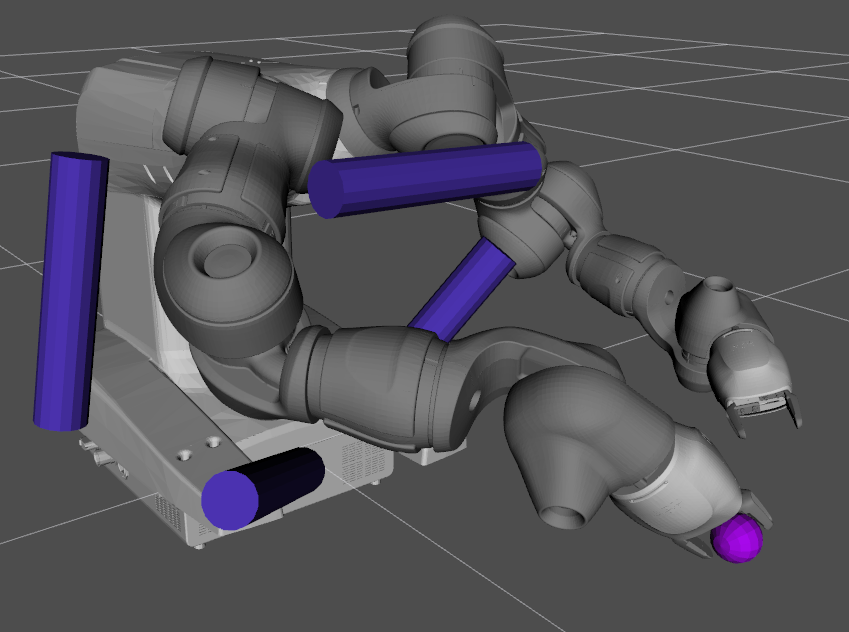}
\includegraphics[height=.30\columnwidth]{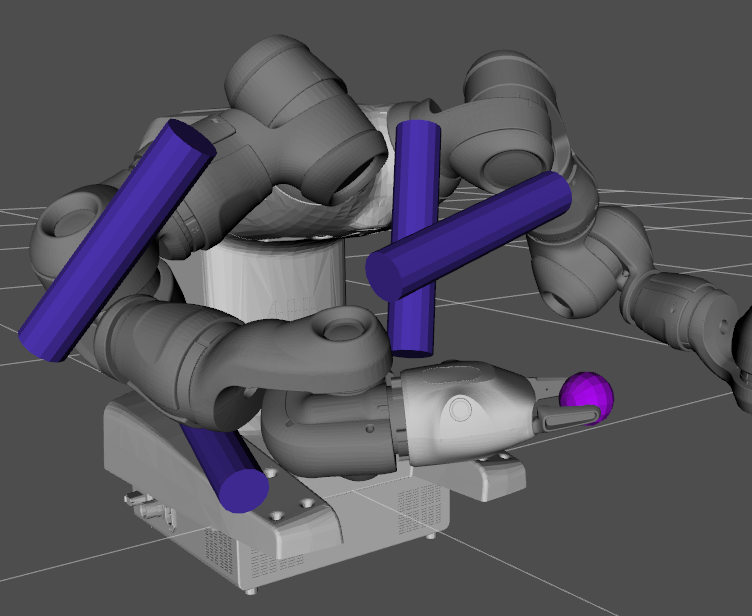}
\end{center}
\caption{The {\bf left three images} shows three of the four obstacle environments and representative start configurations of the retraction experiment. The retracted configuration is depicted by the left arm's configuration in these images; the second retraction heuristic described in Section~\ref{sec:longRange} successfully retracts from all 20 trial configurations in these environments. The simpler heuristic successfully solves all trials except two from the rightmost world which generally has obstacles closer to the robot. The remaining {\bf right two images} depict two of the three reaching environments and solutions found by the RMP motion generation system.}
\label{fig:experimentalEnvs}
\end{figure*}

In practice, it is common to transform behaviors described as dynamical systems on multiple task spaces (especially the end-effector space) into the C-space using pseudoinverses and to combine the resulting policies by superimposing them. Centralized quadratic programming offers a clean framework for combining many such policies (such as in operational space control and its generalizations), but pseudoinverses are still quite common due to their modularity as building blocks for system design. In this section, we show that we can both achieve very strong performance (the final result is theoretically equivalent to running a centralized quadratic program) while maintaining the modularity of pseudoinverses and policy superposition but using RMPs. 

As a baseline, we replace the metric of each pullback policy by equivalently scaled uninformative metric of the form $\beta\I$ to represent the best-scaled pseudoinverse solution. We experimented with a collection of choices for $\beta$, including scalings which would induce uninformative metrics with L2 or L1 norms equivalent to the original metrics, but found choosing $\beta$ to be the maximum eigenvalue of the original metric (matching the L$_\infty$ norm) led to the best performance since it correctly reflected the task strength requirements. In practice, we observed that these uninformative metrics induced substantial competition between the different controllers. In particular, they overpowered the controllers defined directly in the C-space designed to stabilize the system. We, therefore, compared to a progression of such baselines with increasing weight on these C-space controllers. 

Side-by-side comparisons are shown in the supplementary video. For this experiment, we generated 3 cluttered environments of 4 cylindrical obstacles each as depicted in Figure~\ref{fig:experimentalEnvs}, and chose a distribution of target reaching points on the opposite side of the obstacles for the robot to reach to.

In general, as we increase those weights the system becomes more stable and drifts less through null spaces, but increasingly has more difficulty achieving the task. By removing the information encoded in the metric spectrum, even when keeping the overall weightings of the controllers consistent, the controllers no longer have the information necessary to effectively trade-off with one another. As a result they clash or fight with each other leading substantial degradation in performance. This is especially problematic as the number of controllers and distinct task spaces increases. In this case, we use up to 150 controllers in our experiments, most defined in their own task space, and all competing for the resources of just 7 degrees of freedom in the arm.

In some cases, fighting resulted in catastrophic events, such as collision with obstacles. Such collisions can be seen periodically for each of the of the weight settings of the C-space controllers. RMPs successfully solved all of these reaching tasks generating smooth, predictable, and natural motion for each.



\subsection{Retract heuristics for long-range navigation}
\label{sec:experimentsRetraction}

In these experiments, we again generated 4 more cluttered environments with 4 cylindrical obstacles each, manually chose 4 to 6 configurations for each with the robot reaching through, around, between, etc. the objects, and applied the two retract heuristics described in Section~\ref{sec:longRange} to move from the reaching configuration (entwined with the obstacles) back to a retracted ``ready'' configuration.

As noted above in Section~\ref{sec:longRange}, splicing two retraction behaviors together, one played forward and the other played backward implements a long-range guiding policy, since they both move to a common configuration. We do not explicitly show that here, but these heuristics in practice can be easily used to either generate full behaviors that can be followed or as a well-informed technique for seeding motion optimization. We did explicitly implement the IK-guided long-range navigation policy described in Section~\ref{sec:longRange}. This navigation policy was very successful for configuration-to-configuration planning in worlds and problems where the retract heuristics were successful. In general, as long as the elbow is unblocked, these heuristics are quite robust.

The supplementary video shows results of this experiment as well as examples of this long-range navigation policy on a physical Baxter platform in a cluttered world (represented as occupancies found using a depth camera) in an integrated system demoing picking objects from the surface of a table and putting them into a container below the table.

\section{Conclusion}

The RMP framework is a general framework for combining motion policies in a geometrically consistent way. It combines the modularity of pseudoinverses and dynamical system motion representations with the optimality of operational space control techniques. We have found this framework to be invaluable in designing stable and consistent behaviors that combine many different contributions, from many local obstacle controllers to sophisticated continuous motion optimizers that add intelligent coordination and anticipatory behavior. These controllers have been used across a number of single- and dual-arm robotic platforms, three of which we show in the supplementary video. As a side-effect of optimality, we observe that the behavior of the same controllers is surprisingly consistent from robot to robot even without substantial retuning, aside from adjusting for differing length scales. The success of simple heuristics for long-range navigation suggests that more complex heuristics leveraging data-driven pattern recognition approaches such as deep learning is an interesting direction for future work. We have integrated visual feedback for reactive control; as future work also plan to explore designing more dexterous manipulation behaviors leveraging other sensor modalities for reactive feedback within the RMP framework apart from vision, such as touch and force feedback.

{\small
\bibliographystyle{plainnat}
\bibliography{refs}
}



\appendix

\section{Definitions and notation}

Let $\mathcal{C}$ be the C-space parameterized by $\q\in\R^d$ where $d$ is the
C-space
dimension. We typically identify parameter vectors with
configurations and denote $\q\in\mathcal{C}$. A task space $\calX$ is defined
by map $\phi:\mathcal{C}\rightarrow\mathcal{X}$. Notationally, we write $\x =
\phi(\q)$. In all cases, we assume $\calX$ is a smooth manifold. We use
dot-notation for time-derivatives, denoting C-space velocities and
accelerations as $\qd$ and $\qdd$, respectively. Similarly, denoting the map's Jacobian as
$\frac{\partial \phi}{\partial\q} = \Jp$, we have $\xd = \Jp\qd$ and $\xdd =
\Jp\qdd + \Jdp\qd$.

\subsection{A note on generalized inverses}

$\dagger$ denotes the generalized inverse. Let $\A\in\R^{m\times n}$ be an arbitrary $m\times n$ matrix. Every matrix has an SVD $\A = \U\mS\V^T$ with $\U\in\R^{m\times m}$, $\V\in\R^{n\times n}$, and diagonal $\mS\in\R^{m\times n}$. Not all diagonal entries in $\mS$ need be non-zero in general. Let $k\leq m,n$ denote the number of nonzero diagonal entries in $\mS$. The SVD can be written
\begin{align}
  \A = 
  \left[
    \begin{array}{cc}
      | & | \\
      \U_\paral & \U_\perp \\
      | & |
    \end{array}
  \right]
  \left[
    \begin{array}{cc}
      \wt{\mS} & \zero \\
      \zero & \zero
    \end{array}
  \right]
  \Bigg[
    \begin{array}{rcl}
      \mbox{---} & \V_\paral^T & \mbox{---} \\
      \mbox{---} & \V_\perp^T & \mbox{---} \\
    \end{array}
  \Bigg]
  = \U_\paral\ \wt{\mS} \V_\paral^T.
\end{align}
where $\U_\paral\in\R^{m\times k}$, $\V_\paral\in\R^{k\times n}$, and
$\mathrm{diag}\big(\wt{\mS}\big)_i > 0$ for all $i=1,\ldots,k$. Note that $k$
is the rank of $\A$.

\newcommand{\mP}{\mathbf{P}}

Denote the column and row spaces of $\A$ by $\mathrm{col}(\A)$ and $\mathrm{row}(\A)$, respectively, and denote their corresponding projectors as $\mP_{\mathrm{col}(\A)}[\cdot]$ and $\mP_{\mathrm{row}(\A)}[\cdot]$ such that $\mP_{\mathrm{col}(\A)}[\x] \in \mathrm{col}(\A)$ for all $\x\in\R^m$ and $\mP_{\mathrm{row}(\A)}[\y] \in \mathrm{row}(\A)$ for all $\y\in\R^n$. We define the \textit{generalized inverse} as 
\begin{align}
  \A^\dagger = \V_\paral\ \wt{\mS}^{-1}\U_\paral^T \in \R^{n\times m}
\end{align}
so that 
\begin{align}
&\A^{\dagger}\A = \big(\V_\paral\ \wt{\mS}^{-1}\U_\paral^T\big)\big(\U_\paral\ \wt{\mS} \V_\paral^T\big) = \V_\paral\V_\paral^T = \mP_{\mathrm{col}(\A)}[\cdot] \\
&\A\A^{\dagger} = \big(\U_\paral\ \wt{\mS} \V_\paral^T\big)\big(\V_\paral\ \wt{\mS}^{-1}\U_\paral^T\big) = \U_\paral\U_\paral^T = \mP_{\mathrm{row}(\A)}[\cdot].
\end{align}

\subsection{Riemannian Motion Policies}

A {\it motion policy} on a space $\mathcal{X}$ is a smooth nonlinear
time-varying second-order differential equation denoted $\xdd = \f(t, \x,
\xd)$. For notational convenience, we often suppress the time dependence or
even $\x,\xd$ when the context is clear. A {\it Riemannian Motion Policy}
(RMP), denoted $(\f, \G)_{\mathcal{X}}$, is a motion policy $\f$ paired with a
velocity dependent Riemannian metric $\G(\x, \xd)$ defined on the tangent
bundle $T\calX$.
Instantaneously (i.e. for a
given moment in time $t$), we often use the shorthand \begin{align} (\f,
\G)_{\mathcal{X}}\big|_{(\x, \xd)} = (\xdd, \A)_\calX, \end{align} where the
left hand side denotes evaluation at $(\x, \xd)$, and on the right hand side
$\xdd = \f(\x, \xd)$, and $\A = \G(\x, \xd)$. We generally allow the metric to
be degenerate, parameterized as a symmetric semi-positive definite matrix (a
pseudo-Riemannian metric).

Let $\calX_i\ i=1,\ldots,N$ be task spaces with task maps $\x_i = \phi_i(\q)$, and let $(\f_i, \G_i)_{\calX_i}$ be $N$ RMPs with instantaneous evaluations $(\f_i, \G_i)_{\calX_i}|_{(\q, \qd)} = (\xdd_i, \A_i)_{\calX_i}$. The operational space control formalism defines the optimal combined action instantaneously as
\begin{align} \label{eqn:OperationalSpaceControl}
\qdd^* &= \argmin_{\qdd} \sum_{i=1}^N\|\xdd_i - (\J_i\qdd + \Jd_i\qd) \|_{\A_i}^2 \\
&= \argmin_{\qdd} \sum_{i=1}^N\|\xdd_i' - \J_i\qdd \|_{\A_i}^2,
\end{align}
where $\xdd_i' = \xdd_i - \Jd_i\qd$. Some practitioners use the
approximation $\xdd_i' = \xdd_i$ ignoring the second-order (Coriolis) effects
of the nonlinear task map. We adopt that approximation here.

\section{Algebra of RMPs}

Here we define an algebra on RMPs by decomposing the solution to Equation~\ref{eqn:OperationalSpaceControl} into a collection of modular tools that define simple operators on RMPs analogous to pseudoinverses and superposition of differential equations. These tools enable the modular design, transformation, and combination of RMPs while maintaining optimality properties under Equation~\ref{eqn:OperationalSpaceControl}, similar in nature to the least-squares optimality properties of the pseudoinverse, which has become an indispensable tool in linear algebra.

\subsection{Pulling and pushing through maps}

Let $\phi:\mathcal{C}\rightarrow\mathcal{X}$ be a task map. The {\it pullback}
of a metric $\G$ on $\mathcal{X}$ through $\phi$ to $\mathcal{C}$ is $\B =
\J^T\G\J$, where $\J = \frac{\partial \phi}{\partial \q}$. Likewise, if
$(\f,\G)$ is an RMP defined on $\mathcal{X}$, its pullback through $\phi$ to
$\mathcal{C}$ is
\begin{align} \label{eqn:Pullback}
\mathrm{pull}_\phi(\f,\G)_\calX =
  \left(\left(\J^T\G\J\right)^{\dagger}\J^T\G\f,\ \J^T\G\J\right)_{\calC} =
  \left(\B^{\dagger}\vv,\ \B\right)_{\calC},
\end{align}
where $\B = \J^T\G\J$ is the pullback metric and $\vv = \J^T\G\f$ is the
(co-vector) Jacobian-transpose transformation of the vector field to $\phi$'s
domain. The rightmost expression emphasizes the similarity to natural gradient
operators common in machine learning.
More explicitly, $\lambda = \G\f$ is a ``force'' if
$\f$ is an acceleration, and if $\lambda$ came from a potential function,
we would have $\lambda = \nabla_\x\psi(\x)$. Then if $\x = \phi(\q)$, we have
$\nabla_\q \psi(\phi(\q)) = \J^T\nabla_\x\psi$ and the natural gradient under
the pullback metric would be $\left(\J^T\G\J\right)^\dagger\J^T\nabla_\x\psi =
\left(\J^T\G\J\right)^\dagger\J^T\G\f$.

We can analogously define a {\it pushforward} RMP from the domain of a task map
$\x = \phi(\q)$ to its range as
\begin{align} \label{eqn:Pushforward}
  \mathrm{push}^{\phi}(\f, \G)_\calC
  = \Big(\J\f + \Jd\qd,\ \J^\dagger\G(\J^\dagger)^{T}\Big)_\calX
\end{align}
where we treat the second-order terms $\Jd\qd$ explicitly for 
completeness.\footnote{Note that this definition differs somewhat from the 
common definition of pushforward is Riemannian geometry, but we use it in this 
way here because if it's relationship to the RMP pullback operator. Specifically, if 
$\phi$ is invertible, the pushforward and pullbck operations for RMPs are inverses
of each other.}


\subsection{Combination on a given task space}

We define the combination of multiple RMPs on the same space to be their metric-weighted average. Let $(\f_i,\G_i)_{\mathcal{X}}$ be $N$ RMPs all defined on the same domain. Then we define
\begin{align} \label{eqn:Combination}
(\f_c, \G_c) \triangleq \sum_{i=1}^N (\f_i, \G_i) 
&=\left(\left(\sum_i \G_i\right)^{\dagger}\sum_i\G_i\f_i,\ \sum_i \G_i\right)_{\mathcal{X}} \\
&= \left(\G_c^{\dagger}\sum_i\G_i\f_i,\ \G_c\right)_{\mathcal{X}},
\end{align}
where $\G_c = \sum_i \G_i$. If $\G_i = w_i\I$ for $w_i\in\R_+$, this reduces to a traditional weighted average.

\subsection{The unresolved form}

The {\it unresolved} form of an RMP $(\f, \G)_\calX$ is an equivalence class of
pairs of $\wt{\f}(\x, \xd)$ and $\G(\x, \xd)$ (denoted $[\wt{\f},\G]_\calX$) defined as
\begin{align} \label{eqn:UnresolvedForm}
  \big[(\f, \G)_\calX\big] 
  = \big\{\ [\wt{\f},\G]_\calX \ \ \big|\ \ \G^{\dagger}\wt{\f} = \f \ \big\}.
\end{align}
The relationship between the resolved and unresolved forms is analogous to 
the moment and natural parameterizations of Gaussian distributions, 
respectively.\footnote{
The equivalence class here acknowledges that there may be a nullspace of
$\G^\dagger$. As in $\wt{\f} = \wt{\f}_0 + \wt{\f}_\perp$ for any
$\wt{\f}_\perp\in\mathrm{null}(\G^\dagger)$. A simple case is when $\wt{\f} =
\J^T\xdd^d$ and $\G = \J^T\J$. Any $\wt{\f} + \wt{\f}_\perp = \J^T\xdd^d +
\wt{\f}_\perp$ is equivalent here if $\wt{\f}_\perp \in \mathrm{null}(\J^T\J) =
\mathrm{null}(\J)$. In a sense, the
unresolved form works in ``force'' space rather than ``acceleration''
space, analogous to placing forces $\lambda = \A\xdd^d$ in task space,
propagating them into the c-space using $\wt{\f} = \J^T\lambda$ along with
generalized mass matrix $\M = \J^T\A\J$, then solving for the accelerations
$\qdd = \M^\dagger \J^T\lambda = \big(\J^T\A\J\big)^\dagger\J^T\A\xdd^d$ (its
\emph{resolution}). This 
``unresolved'', or force, form is convenient both computationally and in proofs.}

Any element of the equivalent class can be used as a representative
parameterization. {\it Resolution} of an unresolved form to its equivalent {\it
resolved} form is denoted 
\begin{align}
  \big([\wt{\f}, \G]_\calX\big) 
  = \big(\G^{\dagger}\wt{\f},\ \G\big)_\calX.
\end{align}
The resolution of an unresolved form is unique.
It's straightforward to show that $\Big(\big[(\f, \G)_\calX\big]\Big) = (\f, \G)_\calX$.
Operators on the unresolved form become:

\noindent {\bf Pullback.} Given $[\wt{\f}, \G]_\calX$ and $\phi:\calC\rightarrow\calX$ with $\frac{\partial \phi}{\partial \q} = \Jp$:
\begin{align}
  \mathrm{pull}_\phi [\wt{\f}, \G]_\calX 
  = \big[\Jp^T\G\f,\;\Jp^T\G\Jp\big]_\calC.
\end{align}
Note $\Big(\mathrm{pull}_\phi [\wt{\f}, \G]_\calX\Big) = \Big(\big(\Jp^T\G\Jp\big)^{\dagger}\Jp^T\G\f,\;\Jp^T\G\Jp\Big)_\calC$, reproducing Equation~\ref{eqn:Pullback}.

\noindent {\bf Combine.} Given $[\wt{\f}_i, \G_i]_{\calX}$ for $i=1,\ldots,N$:
\begin{align} \label{eqn:UnresolvedCombination}
  \sum_{i=1}^N[\wt{\f}_i, \G_i]_{\calX} 
  = \left[\sum_i\G_i\f_i,\;\sum_i\G_i\right]_\calX.
\end{align}
Again, $\Big(\big[\sum_i\G_i\f_i,\;\sum_i\G_i\big]_\calX\Big) 
= \Big(\big(\sum_i\G_i\big)^{\dagger}\sum_i\G_i\f_i,\;\sum_i\G_i\Big)_\calX$, reproducing Equation~\ref{eqn:Combination}.

The unresolved form is equivalent to the resolved form and often more
computationally efficient and more convenient in proofs.

%
%
\subsection{Associativity, independence of computational path, and optimality}

Suppose we have a tree of task maps with each node representing a task space,
and links representing task maps connecting the spaces. Then if we have RMPs at each of the
task spaces, the question of whether the order with which we pull them back or
combine them matters is paramount. 
Fortunately, we can show that 
this independence 
of computational path follows as a straightforward corollary to the commutativity
and associativity of RMP summation and the linearity of the RMP pullback operator
with respect to the summation (all of which we discuss here).
Similarly, optimality 
can be shown by demonstrating that one particular computational path's result
is straightforwardly equivalent to the solution to a least-squares objective.

\begin{lemma}[Commutativity of summation] Let 
$\mathcal{R}_1 = (\f_1, \A_1)_\mathcal{X}$ and 
$\mathcal{R}_2 = (\f_2, \A_2)_\mathcal{X}$ be RMPs on $\mathcal{X}$.
Then $\mathcal{R}_1 + \mathcal{R}_2 = \mathcal{R}_2 + \mathcal{R}_1$.
\end{lemma}
\begin{proof} The proof is a straightforward computation in the unresolved
form. 
\begin{align}
\mathcal{R}_1 &+ \mathcal{R}_2 
\\&= [\A_1\f_1, \A_1]_\mathcal{X} + [\A_2\f_2, \A_2]_\mathcal{X}
= [\A_1\f_1 + \A_2\f_2, \A_1 + \A_2]_\mathcal{X}
\\&= [\A_2\f_2 + \A_1\f_1, \A_2 + \A_1]_\mathcal{X}
= [\A_2\f_2, \A_2]_\mathcal{X} + [\A_1\f_1, \A_1]_\mathcal{X}
\\&= \mathcal{R}_2 + \mathcal{R}_1.
\end{align}
\end{proof}

\begin{lemma}[Associativity of summation] Let 
$\mathcal{R}_1 = (\f_1, \A_1)_\mathcal{X}$,
$\mathcal{R}_2 = (\f_2, \A_2)_\mathcal{X}$, and 
$\mathcal{R}_3 = (\f_3, \A_3)_\mathcal{X}$ 
be RMPs on $\mathcal{X}$.
Then $(\mathcal{R}_1 + \mathcal{R}_2) + \mathcal{R}_3 
= \mathcal{R}_1 + (\mathcal{R}_2 + \mathcal{R}_3)$.
\end{lemma}
\begin{proof} The proof again is a straightforward calculation, so we omit
it here.
\end{proof}

\begin{lemma}[Linearity of pullback] Let $(\f_1, \A_1)_\mathcal{Y}$ and 
$(\f_2, \A_2)_\mathcal{Y}$ be RMPs on $\mathcal{Y}$, and let 
$\phi:\mathcal{X}\rightarrow\mathcal{Y}$ be a differentiable task map.
Then 
\begin{align}
\mathrm{pull}_\phi \big((\f_1, \A_1)_\mathcal{Y} + (\f_2, \A_2)_\mathcal{Y}\big)
=
\mathrm{pull}_\phi (\f_1, \A_1)_\mathcal{Y} + \mathrm{pull}_\phi(\f_2, \A_2)_\mathcal{Y}.
\end{align}
\end{lemma}
\begin{proof}
This is easy to show by calculation in the unresolved form.\\
$\mathrm{pull}_\phi [\f_i, \A_i]_\mathcal{Y} = [\Jp^T\A\f_i,\;\Jp^T\A\Jp]$, so
\begin{align}
\mathrm{pull}_\phi &\big[[\f_1, \A_1]_\mathcal{Y} + [\f_2, \A_2]_\mathcal{Y}\big] \\
&= \mathrm{pull}_\phi \big[\f_1 + \f_2,\ \A_1 + \A_2\big]_\mathcal{Y} \\
&= \big[\Jp^T\A(\f_1 + \f_2),\ \Jp^T(\A_1 + \A_2)\Jp\big]_\mathcal{Y} \\
&= \big[\Jp^T\A\f_1 + \Jp^T\A \f_2,\ \Jp^T\A_1\Jp + \Jp^T\A_2\Jp\big]_\mathcal{Y} \\
&= \big[\Jp^T\A\f_1,\ \Jp^T\A_1\Jp \big]_\mathcal{Y}
   + \big[\Jp^T\A\f_2,\ \Jp^T\A_2\Jp \big]_\mathcal{Y} \\
&= 
\mathrm{pull}_\phi [\f_1, \A_1]_\mathcal{Y} + \mathrm{pull}_\phi[\f_2, \A_2]_\mathcal{Y}.
\end{align}
\end{proof}

\begin{corollary}[Independence of computational path] \label{cor:IndCompPath}
Let $\mathcal{X}_i,
i=0,1,\ldots,N$  be a collection of task space with some tree-structured
topology defined by a collection of maps
$\phi_j:\mathcal{X}_k\rightarrow\mathcal{X}_l$. Let $\mathcal{X}_0$ be the root
of the tree. And suppose $(\f_i,\A_i)_{\mathcal{X}_i}$ are a collection of RMPs
defined at spaces $\mathcal{X}_i$. Then the pullback and combination of
these RMPs recursively to $\mathcal{X}_0$ is independent of computational path.
\end{corollary}
\begin{proof}(Sketch)
If we recursively show that the pullbacks to a given node $\mathcal{X}_i$ are
independent of path, then by the above linearity property, the pullback of 
each of a parent node's children are independent of computational path. 
Similarly, by the associativity of RMPs all orders of combinating 
those nodes are equivalent. By inducation, noting that each leave starts 
out unique as a combined RMP due to associativity and commutativity, the final result at 
the root must be unique.
\end{proof}

\begin{corollary}[Optimality] Using the setup of Corollary~\ref{cor:IndCompPath}, 
the resulting pullback and combination at the root $\mathcal{X}_0$ is optimal with 
respect to the objective
\begin{align} \label{}
    \mathcal{L}(\qdd) = \sum_i \|\xdd_i^d - \J_{\phi_i}\qdd\|_{\A_i}^2,
\end{align}
where $\xdd_i^d = \f_i(\x, \xd)$ and 
$\phi_i:\mathcal{X}_0\rightarrow\mathcal{X}_i$ denotes the composition of task maps
along the unique path from $mathcal{X}_0$ to $\mathcal{X}_i$ through the tree.
\end{corollary}
\begin{proof}(Sketch)
The proof is a simple calculation that shows the Jacobian $\J_{\phi_i}$ is 
the produce of Jacobians along the path, so the solution to optimization 
of quadratic $\mathcal{L}$ (written in its most basic form) is equivalent
to the sum of direct pullbacks of each RMP from the leaves to the root.
\end{proof}

%
%

\section{Equations of behavior}

This section gives explicit equations used to model behavior in an articulated manipulation system.

\subsection{Directionally stretched metrics}

Directionally stretched Hessian calculation:
Let $\gamma\in\R$ be a scalar. Then the $\alpha$-scaled softmax between $\gamma$ and $-\gamma$ (a soft ``V''-shaped function) is
\begin{align}
h_V^\alpha(\gamma) 
&= \frac{1}{\alpha}\log(e^{\alpha\gamma} + e^{-\alpha\gamma}) 
= \gamma + \frac{1}{\alpha}\log(1 + e^{-2\alpha \gamma}).
\end{align}
That last expression is numerically safe for $\gamma \geq 0$, and an analogous expression can be used for $\gamma \leq 0$. Properties of $h_V$ include:
\begin{enumerate} 
\item $h_V^\alpha(0) = \frac{1}{\alpha}\log(2) > 0$,
\item $\lim_{\gamma\rightarrow\infty} h_V^\alpha(\gamma) / \gamma = 1$, and
\item $\frac{dh_V^\alpha}{d\gamma}(0) = 0$.
\end{enumerate}
These mean (2) it behaves like $\gamma$ for large $\gamma$ (where $\alpha$ defines what large means), but it becomes (1) strictly positive at zero by (3) bending up smoothly to be flat at $0$. 

We use $h_V^\alpha$ to define a {\it soft-normalization} function as
\begin{align} \label{eqn:SoftNormalization}
  \xi_\alpha(\vv) = \vv / h_V^\alpha(\|\vv\|).
\end{align}
$\xi_\alpha(\vv)$ approaches $\hat{\vv} = \frac{\vv}{\|\vv\|}$ for larger $\vv$, but approaches zero smoothly as $\vv\rightarrow 0$.

In many cases, we choose metrics to be ``directionally stretched'' in the sense:
\begin{align}
\A_{\mathrm{stretch}}(\vv) = \xi_\alpha(\vv)\xi_\alpha(\vv)^T.
\end{align}
$\A_{\mathrm{stretch}}(\vv)$ behaves like $\hat{\vv}\hat{\vv}^T$ for larger $\vv$, but approaches $\zero$ as $\vv\rightarrow 0$.

The following is a common pattern: 
\begin{align} \label{eqn:DirectionallyStretchedHessian}
&\mH_\f^\beta(\x, \xd) = \beta(\x)\A_{\mathrm{stretch}}\big(\f(\x, \xd)\big) + \big(1-\beta(\x)\big) \I \\
&\A(\x, \xd) = w(\x)\ \mH_\f^\beta(\x, \xd),
\end{align}
where $\xdd^d = \f(\x, \xd)$ is some smooth differential equation, $\beta(\x) \in [0,1]$ is smooth in $\x$, and $w(\x) > 0$ is a smooth weight function. This metric smoothly transitions from a directionally stretched metric stretching along a desired acceleration vector and an uninformed metric as a function of position, while being modulated by the position-based weight function.

\subsection{Default task spaces}

Let $\mT = \bar{\phi}(\q)$ be a total forward kinematics map with
\begin{align}
  \mT = 
  \left[
    \begin{array}{cc}
      \mR(\q) & \mt(\q) \\
      \zero^T & 1
    \end{array}
  \right]
  \ \ \ \mathrm{where} \ \ \ 
  \mR(\q) = 
  \left[
    \begin{array}{ccc}
      | & | & | \\
      \ma_x & \ma_y & \ma_z \\
      | & | & |
    \end{array}
  \right]
\end{align}
mapping $\q$ to a frame parameterized by rotation $\mR(\q)$ and translation
$\mt(\q)$. Denote the rotational axis components of the frame by $\mT_{x} =
\mR_{x} = \ma_x, \mT_{y} = \mR_{y} = \ma_y, \mT_{z} = \mR_{z} = \ma_z$, and
denote the translational component by $\mT_o = \mt$ (denoting the ``origin'').
As shorthand, we also denote $\x = \mt = \phi(\q)$,  
$\ma_x = \phi_x(\q)$, 
$\ma_y = \phi_y(\q)$, and  
$\ma_z = \phi_z(\q)$.
We call these individual axis and translational components the {\it
frame elements}.

Each robot has a collection of relevant frames along its kinematic chain. We
treat each of these as a forward kinematic task space using superscripts to
distinguish them when needed. For instance, the end effector map is $\mT^e =
\bar{\phi}^e(\q)$ and it's frame elements are given by maps 
$\x = \phi^e(\q)$,
$\ma_x = \phi_x^e(\q)$,
$\ma_y = \phi_y^e(\q)$, 
$\ma_z = \phi_z^e(\q)$.

Below we build each behavior generation RMP on generic task spaces. These task spaces
may be abstract, or they may be one of the above mentioned frame elements. For instance, 
by building attractors on each of the frame elements, we can fully control the 
position and rotation of the robot's end-effector. We can even partially constrain
the rotation by choosing a new task space 
$\vv = \phi_{\vv}(\q) = \alpha_x \phi_x + \alpha_y\phi_y + \alpha_z\phi_z$ for some 
$\alpha_{x,y,z} \in \R$ and building an attactor on that. Many additional 
task spaces can be defined as functions of these frame elements in similar ways, 
such as $z = d(x)\in\R$ the one-dimensional space of distances to a surface.

\subsection{Attractor controllers}

Let $\z = \phi(\q)$ be any task space. Attractors toward a given point $\z_0$ in this space are defined as 
\begin{align} \label{eqn:AttractorController}
  \f_{\mathrm{attract}}(\z) = \gamma_p \xi_\alpha(\z_0 - \z) - \gamma_d \zd,
\end{align}
where $\xi_\alpha(\cdot)$ is the soft-normalization function defined in Equation~\ref{eqn:SoftNormalization}.
$\gamma_p > 0$ and $\gamma_d > 0$ are scalar position and damping gains, respectively. $\alpha > 0$ is chosen to define the effective radius of slowdown close to $\z_0$.

With these controllers, we often choose metrics of the form
\begin{align}
  \A_{\mathrm{attract}} 
  = w_{\mathrm{target}}\big(\|\z_0 - \z\|\big)\ \mH_{\f_{\mathrm{attract}}}^{\beta(\z)}(\z, \zd)
\end{align}
where $\beta(\z) = 1-\exp(-\frac{1}{2}\|\z - \z_0\|^2 / \sigma_{\mH}^2)$ with
$\sigma_{\mH} > 0$, and $w_{\mathrm{target}}(s) = \exp\big(-s / \sigma_w\big)$
with $\sigma_w > 0$.


We use Equation~\ref{eqn:AttractorController} to define attractors for each frame element of a complete forward kinematics map to control the frame origin and any subset of axes. Likewise, in many cases we can define an abstract task space $\z = \psi(\x) = \big(\psi \circ \phi^b\big)(\q)$ where $\x = \phi^b(\q)$ is a forward kinematics function mapping to a body point $b$. $\z$ might, for instance, encode geometric aspects of the space so that straight lines curve around obstacles. RMP attractors in $\z$ defined as in Equation~\ref{eqn:AttractorController}, with the above mentioned directionally scaled metric, pull back into the workspace $\x$ to become nonlinear attractors curving along the natural geometry defined by $\psi$.

A concrete example leverages a cylindrical coordinates coordinates map. Specifically, 
\begin{align}
  \psi_{\mathrm{cyl}}(\x) = 
  \left[
    \begin{array}{c}
      \mr(\x) \\
      \theta(\x) \\
      \z(\x)
    \end{array}
  \right],
\end{align}
where $\mr(\x)$ is the radius to the cylinder's axis, $\z(\x)$ is the height along the axis, and $\theta(\x)$ is the angle around the axis in radians. Define $\wt{\psi}_{\mathrm{cyl}}(\x) = \W \psi_{\mathrm{cyl}}(\x)$, where $\W$ is an appropriate diagonal positive definite weighting function, often designed with the weight on $\mr(\x)$ large so that the pullback metrics encourage geodesics to move along cylindrical curves equidistant from the cylinder's axis. Within $\z = \wt{\psi}_{\mathrm{cyl}(\x)}$, attractors defined by Equation~\ref{eqn:AttractorController} will naturally curve around the cylindrical axis of rotation.

\subsection{Collision avoidance controllers}

Let $d:\mathcal{X}\rightarrow \R_+$ be a distance function $d(\x)$. Note that the distance gradient is innately normalized and points away from the obstacle, i.e $\|\nabla d(\x)\| = 1$ and $\nabla d^T(\x_{\mathrm{obs}} -\x) > 0$ where $\x_{\mathrm{obs}}$ is the closest obstacle point ($\x_{\mathrm{obs}} = \x -d(\x) \nabla d(\x)$). 

The obstacle avoidance controller is composed of a repulsive term $\f_{\mathrm{rep}}(\x)$ and a damping term $\f_{\mathrm{damp}}(\x, \xd)$:
\begin{align}
  \f_{\mathrm{obs}}(\x, \xd) = \f_{\mathrm{rep}}(\x) + \f_{\mathrm{damp}}(\x, \xd).
\end{align}
The repulsive term can be written
\begin{align}
  \f_{\mathrm{rep}}(\x) = \alpha_{\mathrm{rep}}(\x)\nabla d(\x),
\end{align}
where $\alpha_{\mathrm{rep}}(\x)$ is a position-based activation function define as
\begin{align}
  \alpha_{\mathrm{rep}}(\x) = \eta_{\mathrm{rep}}\exp\left({-\frac{d(x)}{\nu_{\mathrm{rep}}}}\right)
\end{align}
where $\nu_{\mathrm{rep}} > 0$ is a positive length scale and $\eta_{\mathrm{rep}} > 0$ is a constant gain. 

Let $P_{\mathrm{obs}}(\x)[\xd]$ denote a directionally-scaled projection operator defined as
\begin{align} \label{eqn:ObsVelProj}
  P_{\mathrm{obs}}(\x)[\xd] 
  = \max\big\{0, -\xd^T\nabla d\big\}\Big[\nabla d \nabla d^T\Big] \xd
\end{align}
which projects $\xd$ onto the direction toward the obstacle while scaling it by a factor that vanishes as $\xd$ moves toward the half space $\mathcal{H}_{\mathrm{away}} = \{v\ |\ \nabla d(\x)^Tv \geq 0\}$ orthogonal to or pointing away from the obstacle. Note that $P_{\mathrm{obs}}(\x)[\xd]$ has a continuous derivative in both $\x$ and $\xd$ when $d(\x)$ is second-order differentiable, which can be seen by rewriting it as 
$P_{\mathrm{obs}}(\x)[\xd] = -\big(\xd^T\nabla d\big)_+^2 \nabla d$, where $(\upsilon)_+ = \max\{0, \upsilon\}$. 



Using $P_{\mathrm{obs}}$, we define the damping term as
\begin{align} \label{eqn:CollisionDamping}
  \f_{\mathrm{damp}}(\x, \xd) = \alpha_{\mathrm{damp}}(\x) P_{\mathrm{obs}}(\x)[\xd],
\end{align}
where $\alpha_{\mathrm{damp}}(\x)$ is position-based activation function given by
\begin{align}
  \alpha_{\mathrm{damp}}(\x) = \eta_{\mathrm{damp}}\Big/\left(\frac{d(\x)}{\nu_{\mathrm{damp}}} + \epsilon\right)
\end{align}
with $\eta_{\mathrm{damp}} > 0$ a constant gain. We include $0 < \epsilon \ll 1$ for numerical stability as $s\rightarrow 0$.

The corresponding metric is $\A_{\mathrm{obs}}(\x, \xd) = w_r\big(d(\x)\big)
\mH_{\f_{\mathrm{obs}}}^0(\x, \xd)$ where $\mH$ is given by
Equation~\ref{eqn:DirectionallyStretchedHessian} and $w_r:\R\rightarrow\R$ is a
weight function defining the policy's overall activation, derived as a cubic
spline between 
$(s_0, w_r(s_0), w'_r(s_0)) = (0, 1, 0)$
and
$(s_1, w_r(s_1), w'_r(s_1)) = (r, 0, 0)$.
Specifically,
$w_r(s) = c_2 s^2 + c_1 s + c_0$ with $c_2 = \frac{1}{r^2}, c_1 = -\frac{2}{r}, c_0 = 1$.

Together, $(\f_\mathrm{obs}, \A_{\mathrm{obs}})_\calX$ form an obstacle avoidance RMP.

\subsection{C-space biasing}

Redundancy resolution is implemented as a spring-damper system in the C-space:
\begin{align} \label{eqn:RedundancyResolution}
\f_{\mathrm{res}}(\q, \qd) = \gamma_p (\q_0 - \q) - \gamma_d (\qd_0 - \qd),
\end{align}
where $\gamma_p,\gamma_d\in\R_+$ are gains and $\q_0,\qd_0$ are target
positions and velocities. Often $\qd_0 = \zero$ to implement pure damping. In
practice, this equation typically interacts with the sigmoidal joint limit
handling map and is replaced by
Equation~\ref{eqn:SigmoidScaledRedundancyResolution}.

\subsection{Handling joint limits using pullbacks}

We handle joint limits $q_i \in [\ubar{l}_i, \bar{l}_i]$ using a nonlinear task map mapping from an unconstrained space to the joint limit constrained C-space. The simplest such map is an affine-transformed sigmoid map $\q = \sigma_{\mL}(\uu)$ operating independently per dimension with entries
\begin{align} \label{eqn:JointLimitSigmoid}
  q_i = \sigma_i(u_i) = (\bar{l}_i - \ubar{l}_i)\sigma(u_i) + \ubar{l}
  = \sigma \bar{l}_i + (1-\sigma) \ubar{l}_i,
\end{align}
where $\sigma(u) = 1/(1+e^{-u})$. Given any RMP $(\f, \A)_{\calC}$ defined on the C-space, we pull it back through $\sigma_{\mL}$ and add a simple regulator RMP $(\h, \lambda\I)$, $\lambda>0$, designed to enforce $\uu\in\mathcal{B}$, where $\mathcal{B}$ is some zero centered ball. For any C-space RMP system $(\f, \A)_{\calC}$ with bounded energy, there exists such a bounding ball $\mathcal{B}$ for each $\h(\uu, \ud) = \gamma_p (\zero - \uu) - \gamma_d\ud$ with $\gamma_p, \gamma_d > 0$, so there is substantial flexibility in the choice of $\h$. We will denote the diagonal Jacobian as $\frac{\partial \sigma_{\mL}}{\partial \uu} = \D_\sigma$.

In practice, we use a velocity-dependent differentiable map $\q = \wt{\sigma}_\mL(\uu, \ud)$ with Jacobian $\wt{\D}_\sigma(\uu,\ud) = \frac{\partial \wt{\sigma}_\mL}{\partial\uu}$. Let 
$\sigma_i = 1/(1 + e^{-u_i})$ and $\alpha_i =1/(1+e^{-c\dot{q}})$ with $c\in\R_+$, then
denoting $d_{ii} = \mathrm{diag}[\D_\sigma]_i$ and $\wt{d}_{ii} = \mathrm{diag}[\wt{\D}_\sigma]_i$ we define
\begin{align} \label{eqn:VelcityDependentSigmoidJacobian}
\wt{d}_{ii} = 
    \sigma_i \Big(\alpha_i d_{ii} + (1-\alpha_i) 1\Big) 
    + (1-\sigma_i) \Big((1-\alpha_i) d_{ii} + \alpha_i 1\Big).
\end{align}
Intuitively, we can interpret $\sigma_i$ as ``if $q_i$ is close to $\bar{l}$'' and $\alpha_i$ as ``if $\dot{q}_i > 0$'' and similar but opposite logic for $1-\sigma_i$ and $1-\alpha_i$ with $\ubar{l}$ and negative velocities. This equation turns $d_{ii}$ on and off based on proximity to a joint limit and velocity: if $q_i$ is close to a joint limit and heading toward it, use $d_{ii}$ otherwise use $1$ ($=\mathrm{diag}[\I]_i$).


We show here that the pullback and joint limit regulation can be implemented using a simple augmentation to the original problem, while remaining in $\q$ and not requiring an explicit evaluation of $\wt{\sigma}_\mL$. The resulting primary operation involves down-weighing the columns of the Jacobian based on joint limit proximity.
We derive this result assuming $\A$ is full rank; the reduced rank argument is similar. For notational simplicity we use $\sigma_\mL$, but the result holds for $\wt{\sigma}_\mL$, which is what we use in practice. 

Using the RMP pullback operation from Equation~\ref{eqn:Pullback} we get
\begin{align}
\mathrm{pull}_{\sigma_\mL}\big[(\f, \A)_\calC\big] 
&= \Big(\big(\D_\sigma\A\D_\sigma\big)^{-1}\D_\sigma\A\f,\ \D_\sigma\A\D_\sigma\Big)_\calU\\
&= \Big(\D_\sigma^{-1}\f,\ \D_\sigma\A\D_\sigma\Big)_\calU.
\end{align}
Combining with $(\h, \lambda\I)_\calU$ using Equation~\ref{eqn:Combination} gives
\begin{align}
&\big(\D_\sigma^{-1}\f,\ \D_\sigma\A\D_\sigma\big)_\calU 
+ \big(\h, \lambda\I\big)_\calU \\
&\ \ \ = 
\bigg(
  \Big(\D_\sigma\A\D_\sigma + \lambda\I\Big)^{-1} 
    \big(\D_\sigma\A\f + \lambda\h\big),\ 
  \D_\sigma\A\D_\sigma + \lambda\I
\bigg)_\calU.
\end{align}
Pushing the result forward through $\sigma_{\mL}$ to $\calC$ gives the final result as
\begin{align} \label{eqn:JointLimitCorrected}
&\mathrm{push}^{\sigma_\mL} \Big[
  \mathrm{pull}_{\sigma_\mL}\big[(\f, \A)_\calC\big]
  + \big(\h, \lambda\I\big)_\calU
\Big] = \Big(\f_{\mL}(\q, \qd),\ \G_{\mL}(\q, \qd)\Big)_\calC \\\nonumber
&\ \ \ =
\bigg(
  \D_\sigma \Big(\D_\sigma\A\D_\sigma + \lambda\I\Big)^{-1} 
    \big(\D_\sigma\A\f + \lambda\h\big),
  \A + \lambda\D_\sigma^{-2}
\bigg)_\calC.
\end{align}
Note that this metric $\A + \lambda\D_\sigma^{-2}$ becomes large along for
dimensions close to joint limits.

If $\A = \sum_{i=1}^N \J_i\A_i\J_i$ is a combined collection of pullback metrics, then 
\begin{align}
\D_\sigma\A\D_\sigma = \sum_i \big(\D_\sigma\J_i^T\big)\; \A_i\; \big(\J_i\D_\sigma\big)
= \sum_i \wt{\J}_i^T \A_i \wt{\J}_i,
\end{align}
where $\wt{\J}_i = \J_i\D_\sigma$. Likewise, if 
$\f = \left(\sum_i\J_i^T\A_i\J_i\right)^\dagger \sum_i \J_i^T\A_i\xdd_i
= \A^\dagger \sum_i \J_i^T\A_i\xdd_i$,
we get
\begin{align}
\D_\sigma\A\f = \sum_i \big(\D_\sigma\J_i^T)\A_i\xdd_i
= \sum_i \wt{\J}_i^T\A_i\xdd_i.
\end{align}
Therefore, the differential equation portion of Equation~\ref{eqn:JointLimitCorrected}, denoted $\f_{\mathrm{\mL}}(\q, \qd)$, reduces to
\begin{align}
\f_{\mathrm{\mL}}(\q, \qd) =
\D_\sigma \bigg(\sum_i \wt{\J}_i^T \A_i \wt{\J}_i + \lambda\I\bigg)^{-1} 
    \Big(\sum_i \wt{\J}_i^T\A_i\xdd_i + \lambda\h\Big).
\end{align}
Note that this expression is equivalent to the joint limit free combined task
space RMPs, except with the following additional operations:
\begin{enumerate}
\item Scale each Jacobian by $\D_\sigma$ to get $\wt{\J} = \J\D_\sigma$.
\item Add a regularizer $\lambda\I$ to the final pullback metric.
\item Add $\big(\h(\sigma^{-1}(\q), \D_\sigma^{-1}\qd),\ \lambda\I\big)_\calC$ as a new C-space RMP.
\end{enumerate}
Note that Step (1) plays a significant role:
\begin{align}
\wt{\J} =
  \left[
  \begin{array}{cccc}
    \hspace{25pt}| & \hspace{25pt}| & & \hspace{25pt}| \\
    \sigma_1'(u_1)\J_{:1} & \sigma_2'(u_2)\J_{:2} & \cdots & \sigma_d'(u_d)\J_{:d} \\
    \hspace{25pt}| & \hspace{25pt}| & & \hspace{25pt}|
  \end{array}
  \right],
\end{align}
where $\J_{:j}$ is the $j^{\mathrm{th}}$ column of $\J$.
As $\sigma_j'$ vanishes (near limits), the Jacobian columns are weighed down
toward zero to reduce the final policy's dependency on them.

Since it is unclear what Equation~\ref{eqn:VelcityDependentSigmoidJacobian} integrates to, we do not have a closed form solution for the corresponding task map, so evaluating $\h(\uu, \ud) = \h(\sigma^{-1}(\q), \D_\sigma^{-1}\qd)$ is hard in general. We, therefore, choose $\h$ to avoid such evaluation:
\begin{align} \label{eqn:SigmoidScaledRedundancyResolution}
  \h\big(\sigma^{-1}(\q), \D_\sigma^{-1}\qd\big) 
  = \D_\sigma^{-1}\Big(\gamma_p\big(\q_0 - \q\big) - \gamma_d\qd\Big),
\end{align}
where $\gamma_p,\gamma_d\in\R_+$. Near joint limits, the accelerations of this simple spring-damper are amplified. This expression replaces $\f_{\mathrm{res}}(\q, \qd)$ given in Equation~\ref{eqn:RedundancyResolution}.


%
%
%

\section{Integration with motion optimization}

RMPs can also be computed using more complex planners and motion optimizers.
We present here the basic theoretical reduction. Note that motion optimization
is a subclass of optimal control \cite{OptimalControlEstimationStengel94} 
problems (although in practice the solution 
techniques sometimes differ), so we present the result here in the 
broader context of optimal control.

Let $\s$ be a state, such as $\s = (\q, \qd)$. Given a nonlinear dynamics 
function $\s_{t+1} = \f(\s_t, \ma_t)$, the motion optimization 
problem is
\begin{align} \label{eqn:MotionOpt}
    &\min_{\ma_{1:T}} \sum_{t=1}^T c_t(\s_t, \ma_t) + V_{T+1}(\s_{T+1}) \\
    &\mathrm{s.t.}\ \ \s_{t+1} = \f(\s_t, \ma_t) \\
    &\ \ \ \ \ \ \g_t(\s_t,\ma_t) \leq \zero, \\
    &\ \ \ \ \ \ \h_t(\s_t,\ma_t) = \zero,
\end{align}
where $\g_t$, $\h_t$ are inequality and equality constraints, respectively.
and $t=1,\ldots,T$ in all constraints (we drop terminal constraints for uniformity, 
although those can be easily incorporated as well as functions of state alone).

Around a local minimum $\ma_{1:T}^*$ we can form an unconstrained proxy objective
whose local minimum satisfies the KKT constraints of the Problem~\ref{eqn:MotionOpt}
in a number of ways. For instance, Augmented Lagrangian
\cite{NocedalWright2006,14-toussaint-AugLag} is such a method.
Let 
\begin{align} \label{eqn:MotionOptUnconstrained}
    &\min_{\ma_{1:T}} \sum c_t^{a}(\s_t, \ma_t) + V_{T+1}(\s_{T+1}) \\
    &\mathrm{s.t.}\ \ \s_{t+1} = \f(\s_t, \ma_t),
\end{align}
denote such an augmented objective where we explicitly include the dynamics
constraints as separate constraints.

Unconstrained optimal control problems of the form in 
Equation~\ref{eqn:MotionOptUnconstrained} are ammenable transformation to local linear
policies using DDP \cite{OptimalControlEstimationStengel94}. We use a similar transformation here
to produce a series of Q-functions representing the optimal solution.
Define $Q_t(\s_t,\ma_t) = c_t^a(\s_t,\ma_t) + V_{t+1}(\f(\s_t,\ma_t))$
with the recurrence:
\begin{align} \label{eqn:DDPRecurrence}
    V_t(\s_t) 
    &= \min_{\ma_t} Q(\s_t, \ma_t) \\
    &= \min_{\ma_t} \Big\{c_t^a(\s_t,\ma_t) + V_{t+1}(\f(\s_t,\ma_t))\Big\}.
\end{align}
And denoting $\z_t = (\s_t,\ma_t)$, let 
$\wt{Q}(\z_t) = \frac{1}{2}(\z-\z_t)^T\nabla^2Q(\z-\z_t) + \nabla Q^T(\z-\z_t) + c$ 
denote the second-order approximation of $Q$ for some $c\in\R$. Then 
each minimization over $\ma_t$ can be solved analytically and the resulting 
$\wt{V}_t(\s_t)$ is quadratic. Moreover, we can solve for 
$\ma_t^* = \pi_t(\s_t) = \argmin_{\ma_t}\wt{Q}_t(\s_t, \ma_t)$ in closed 
form, and $\pi_t$ is linear.

Therefore, fixing $\s_t$, we can write
\begin{align}
    \wt{Q}(\ma_t|\s_t) = \frac{1}{2}\|\pi_t^*(\s_t) - \ma_t\|_{\nabla^2_{\ma_t}\wt{Q}}^2.
\end{align}
Thus, deviations from the optimal policy can be summarized as an RMP
of the form $\big(\pi_t^*(\s_t), \nabla^2_{\ma_t}\wt{Q}\big)_{\mathcal{C}}$.
We call such a stream of RMPs formed of 
time-varying affine policies $\pi_t^*$ with metrics $\nabla^2_{\ma_t}\wt{Q}$ 
a \emph{reduction} 
of motion optimization to RMP.

\end{document}


\title{Riemannian Motion Policies -- Appendix}
\date{}

\maketitle

\section{Derivations}

The algebra of RMPs follows the mathematics of quadratic functions, which is used in Riemannian geometry. 

\subsection{Definitions}

Let $\calQ = \Real^d$ be the configuration space, and let $\calX_1, \calX_2, \ldots$ be a set of task spaces, where $\calX_i = \Real^{k_i}$.  Let $\q(t), \qd(t), \qdd(t) \in \calQ$ be the position, velocity, and acceleration of the robot at time $t$.  Similarly, let $\x_i(t), \xd_i(t), \xdd_i(t) \in \calX_i$ be the position, velocity, and acceleration of the $i^{\text{th}}$ task variable at time $t$.  We assume second-order dynamical systems, so that $\qdd(t)=\f(\q(t),\qd(t))$, where $\f$ is some arbitrary function, and similarly for $\xdd(t)$.

The $i^{\text{th}}$ differentiable task map $\phi_i:\Real^d\rightarrow\Real^{k_i}$ relates the configuration space to the $i^{\text{th}}$ task space, so that $\x_i = \phi_i(\q)$.  In the following, we drop the subscript for simplicity, and we often drop the explicit dependence upon time.

Let us denote the Jacobian of a task map $\phi$ as
\begin{align}
	\Jp \equiv \frac{\partial\phi}{\partial\q} \in \Real^{k \times d}.
\end{align}
The task space velocities and accelerations are then given by
\begin{align}
	\xd &= \frac{d}{dt}\phi(\q) = \Jp\qd \label{eq:appxdeqjqd} \\
    \xdd &= \frac{d^2}{dt^2}\phi(\q) = \Jp\qdd + \Jdp\qd \approx \Jp\qdd, \label{eq:appxddeqjqdd} 
\end{align}
where the last approximation drops the term associated with the second-order curvature of $\phi$.

For a given Riemannian metric $\A \in \Real^{k \times k}$, let us represent the inner product of a vector with itself according to the metric as
\begin{align}
\|\x\|^2_\A = \langle \x, \x \rangle = \x^\T \A \x.
\end{align}
If we define
\begin{align}
\B \equiv \J^\T\A\J
\label{eqn:appbjaj}
\end{align}
then from \myeqref{eq:appxdeqjqd}--\eqref{eqn:appbjaj} we have \begin{align}
    \|\xd\|_\A^2 = \xd^\T\A\xd &= \qd(\J^\T\A\J)\qd = \|\qd\|_{\B}^2 \label{xdaqdb} \\
    \|\xdd\|_\A^2 = \xdd^\T\A\xdd &= \qdd(\J^\T\A\J)\qdd = \|\qdd\|_{\B}^2, \label{xddaqddb}
\end{align}
where we drop the subscript on the Jacobian when it is clear from the context.  

\subsection{Riemannian motion policies (RMPs)}

Let us define a Riemannian motion policy (RMP) in some space $\Omega$ as a tuple ${^\Omega}(\f,\A)$, where $\f:\omega,{\dot \omega} \mapsto {\ddot \omega}$, so that 
${\ddot \omega} = \f(\omega,{\dot \omega})$, and where $\omega(t), {\dot \omega}(t), {\ddot \omega}(t) \in \Omega$ are the position, velocity, and acceleration of some variable in the space.

\subsubsection{RMP cost functions}

Let us associate each RMP $\calR={^\Omega}(\f,\A)$ with a cost function $c_\calR$ given by the inner product of the vector fields $\f$ and $\p$ evaluated at some point in $\Omega$, multiplied by 0.5 for convenience:
\begin{align}
c_\calR(\p) \equiv \frac{1}{2} \|\f - \p\|^2_\A.
\end{align}
Intuitively, $c_\calR$ captures the dissimilarity between the two vector fields, according to the metric $\A$.

\subsubsection{RMP addition}

We define the sum of two RMPs $\calR_1={^\Omega}(\f_1,\A_1)$ and $\calR_2={^\Omega}(\f_2,\A_2)$ in some space $\Omega$ as
\begin{align}
\calR_c = \calR_1+\calR_2 = {^\Omega}((\A_1+\A_2)^+(\A_1\f_1+\A_2\f_2),\A_1+\A_2),
\end{align}
that is, $\calR_c={^\Omega}(\f_c,\A_c)$ where
\begin{align}
\f_c &= (\A_1+\A_2)^+(\A_1\f_1+\A_2\f_2) \\
\A_c &= \A_1+\A_2,
\end{align}
where $^+$ is the pseudoinverse.  With this definition, it is easy to see that RMP addition is both commutative:
\begin{align}
\calR_2+\calR_1 &= {^\Omega}((\A_2+\A_1)^+(\A_2\f_2+\A_1\f_1),\A_2+\A_1) \\
&= {^\Omega}((\A_1+\A_2)^+(\A_1\f_1+\A_2\f_2),\A_1+\A_2) \\
&= \calR_1 + \calR_2
\end{align}
and associative:  $(\calR_1+\calR_2)+\calR_3=\calR_1+(\calR_2+\calR_3)$.  The summation of several RMPs is easily obtained by applying the definition repeatedly:
\begin{align}
\sum_i \calR_i &= \calR_1+\calR_2+\cdots \\
&= {^\Omega}((\A_1+\A_2+\cdots)^+(\A_1\f_1+\A_2\f_2+\cdots),\A_1+\A_2+\cdots) \\
&= {^\Omega}\left(\left(\sum_i \A_i\right)^+\left(\sum_i \A_i\f_i\right),\sum_i \A_i\right).
\end{align}
The identity with respect to addition is ${^\Omega}(\bfzero, \I)$, where $\bfzero$ is a vector of all zeros, and $\I$ is the identity matrix.

\subsubsection{Pullback and pushforward of RMPs}

Suppose we have a mapping $\phi: \calQ \rightarrow \calX$ from some space $\calQ$ to another space $\calX$.  Let us define the \emph{pullback} of an RMP ${^\calX}(\f,\A)$ defined in the co-domain $\calX$ to be the following RMP defined in the domain $\calQ$:
\begin{align} \label{eqn:apppullback}
 \mathtt{pull}_\phi\left({^\calX}(\f, \A)\right)= {^\calQ}\left(\left(\J^\T\A\J\right)^{+}\J^\T\A\f,\ \J^\T\A\J\right).
\end{align}
Similarly, let us define the \emph{pushforward} of an RMP $(\h,\B)$ defined on the domain $\calQ$ to be the following RMP defined on the co-domain $\calX$:
\begin{align} \mathtt{push}_\phi\left({^\calQ}(\h, \B)\right) = {^\calX}\left(\J\h, \,\,{(\J^{+})}^\T\B\J^+\right).
\end{align}

\textbf{Simplified case when $\J$ is full rank.}
If $\J$ is full row rank (i.e., the rows of $\J$ are linearly independent), and $\A$ is strictly positive definite (and therefore full rank), then the pullback vector field simplifies to
\begin{align} \label{eqn:apppullbacksimpleshow}
\left(\J^\T\A\J\right)^{+}\J^\T\A\f
 &=
\left(\J^+\A^{-1}(\J^\T)^+\right)\J^\T\A\f = \J^+\f,
\end{align}
in which case the pullback can be written as
\begin{align} \label{eqn:apppullbacksimple}
 \mathtt{pull}_\phi\left({^\calX}(\f, \A)\right)= {^\calQ}\left(\J^+\f,\ \J^\T\A\J\right).
\end{align}
In the case that $\J$ is full row rank, it is easy to see that $\mathtt{pull}$ and $\mathtt{push}$ are inverses of each other.  That is, 
\begin{align} \mathtt{push}_\phi\left(\mathtt{pull}_\phi\left({^\calX}(\f, \A)\right)\right) 
&= {^\calX}\left(\J(\J^+\f), \,\,{(\J^{+})}^\T(\J^\T\A\J)\J^+\right) \\
&= {^\calX}\left(\f, \,\,\A\right) \\
\mathtt{pull}_\phi\left(\mathtt{push}_\phi\left({^\calQ}(\h, \B)\right)\right) 
&= {^\calQ}\left(\J^+(\J\h), \,\,\J^\T((\J^+)^\T\B\J^+)\J\right) \\
&= {^\calQ}\left(\h, \,\,\B\right).
\end{align}

\textbf{Associativity.}
We now show that the pullback and pushforward operators are associative when $\J$ is full row rank.  Let $\z = \phi_1(\q)$ and $\x = \phi_2(\z)$ so that $\x = (\phi_2\circ\phi_1)(\q) = \phi_2(\phi_1(\q))$ is well defined.  Then the Jacobians are given by
\begin{align}
	\J_{\phi_1} &= \frac{\partial{\phi_1}}{\partial\q} \\
	\J_{\phi_2} &= \frac{\partial{\phi_2}}{\partial\z} \\
	\J_{\phi} &= \frac{\partial{\phi}}{\partial\q} = \frac{\partial{\phi_2}}{\partial\z} \cdot \frac{\partial{\phi_1}}{\partial\q} = \J_{\phi_2}\J_{\phi_1}.
\end{align}
Suppose $\calR={^\calX}(\f,\A)$ is an RMP on $\calX$.  Then
\begin{align}
 \mathtt{pull}_{\phi_1}\big(\mathtt{pull}_{\phi_2}(\calR)\big) 
  &= {^\calQ}\left(\J_{\phi_1}^+(\J_{\phi_2}^+\f), \J_{\phi_1}^\T (\J_{\phi_2}^\T\A\J_{\phi_2}) \J_{\phi_1}\right) \\
  &= {^\calQ}\left((\J_{\phi_2}\J_{\phi_1})^+\f), (\J_{\phi_2}\J_{\phi_1})^\T\A(\J_{\phi_2} \J_{\phi_1})\right) \\
  &= {^\calQ}\left(\J_{\phi}^+\f, \J_{\phi}^\T\A\J_{\phi}\right) \\
 &=
 \mathtt{pull}_{\phi}(\calR) 
 \end{align}
Similarly, suppose $\calR'={^\calQ}(\h,\B)$ is an RMP on $\calQ$.  Then
 \begin{align}
 \mathtt{push}_{\phi_2}\big(\mathtt{push}_{\phi_1}(\calR')\big) &= {^\calX}\left(\J_{\phi_2}(\J_{\phi_1}\h), \,\,{(\J_{\phi_2}^{+})}^\T({(\J_{\phi_1}^{+})}^\T\B\J_{\phi_1}^+)\J_{\phi_2}^+\right) \\
&= {^\calX}\left((\J_{\phi_2}\J_{\phi_1})\h, \,\,{((\J_{\phi_2}\J_{\phi_1})^{+})}^\T\B(\J_{\phi_2}\J_{\phi_1})^+\right) \\
&= {^\calX}\left(\Jp\h, \,\,{(\Jp^{+})}^\T\B\Jp^+\right) \\
&= \mathtt{push}_\phi\left(\calR'\right).
\end{align} 

\textbf{Linearity.}  We now show that the pullback and pushforward operators are linear when $\J$ is full row rank.  Let $\calR_1$ and $\calR_2$ be RMPs in some space.  Then
\begin{align}
\mathtt{pull}_{\phi}(\calR_1) &= \left(\J^+\f_1, \J^\T\A_1\J\right) \\
\mathtt{pull}_{\phi}(\calR_2) &= \left(\J^+\f_2, \J^\T\A_2\J\right)
\end{align}
so that 
\begin{align}
\mathtt{pull}_{\phi}(\calR_1) +
\mathtt{pull}_{\phi}(\calR_2) &= \left((\J^\T\A_1\J+\J^\T\A_2\J)^+(\J^\T\A_1\J\J^+\f_1+\J^\T\A_2\J\J^+\f_2), \J^\T\A_1\J+\J^\T\A_2\J\right) \\
&=
\left((\J^\T(\A_1+\A_2)\J)^+(\J^\T(\A_1\f_1+\A_2\f_2)), \J^\T(\A_1+\A_2)\J\right) \\
&=
\left(\J^+(\A_1+\A_2)^+(\J^\T)^+\J^\T(\A_1\f_1+\A_2\f_2), \J^\T(\A_1+\A_2)\J\right) \\
&=
\left(\J^+(\A_1+\A_2)^+(\A_1\f_1+\A_2\f_2), \J^\T(\A_1+\A_2)\J\right) \label{eqn:applinearityproof1}
\end{align}
The sum of the RMPs is given by
\begin{align}
\calR_1 + \calR_2 &=
\left((\A_1+\A_2)^+(\A_1\f_1+\A_2\f_2), \A_1+\A_2\right),
\end{align}
so that
\begin{align}
\mathtt{pull}_{\phi}(\calR_1+\calR_2) &= \left(\J^+(\A_1+\A_2)^+(\A_1\f_1+\A_2\f_2), \J^\T(\A_1+\A_2)\J\right).  \label{eqn:applinearityproof2}
\end{align}
Comparing \myeqref{eqn:applinearityproof1}
with \myeqref{eqn:applinearityproof2}, we see that
\begin{align}
\mathtt{pull}_{\phi}(\calR_1+\calR_2) &=
\mathtt{pull}_{\phi}(\calR_1) +
\mathtt{pull}_{\phi}(\calR_2),
\end{align}
which is what we aimed to prove.  The proof for pushforward is similar.

 

\subsection{The motion generation problem}

Suppose we have a set of task maps $\phi_i$ with associated desired acceleration vector fields $\xdd_i^d$ with Riemannian metrics $\A_i$.  Our goal is to find a motion policy $\f:\q,\qd \mapsto \qdd$ such that
\begin{align}
    \f(\q,\qd) = \arg \min_{\qdd} \sum_i \frac{1}{2}\|\xdd_i^d - \underbrace{\J_{\phi_i}\qdd}_{\xdd_i}\|_{\A_i}^2. \label{eq:appquadraticsum}
\end{align}
That is, we wish to find the second-order dynamical system that minimizes the cost function combining all the desired accelerations while taking into account their associated metrics.  This cost function $c$ is defined so that $\f = \arg \min_{\qdd} c(\qdd)$:
\begin{align}
    c({\qdd}) &= \sum_i \frac{1}{2}\|\xdd_i^d - \underbrace{\J_{\phi_i}\qdd}_{\xdd_i}\|_{\A_i}^2 \\
    &= \sum_i \frac{1}{2}\left(\xdd_i^d - \J_{\phi_i}\qdd\right)^\T\A_i\left(\xdd_i^d - \J_{\phi_i}\qdd\right)\label{eq:appquadraticsumcost2} \\
    &= \frac{1}{2}\sum_i 
(\xdd_i^d)^\T\A_i(\xdd_i^d) 
- \qdd^\T\J_{\phi_i}^\T\A_i\xdd_i^d
  - (\xdd_i^d)^\T\A_i\J_{\phi_i}\qdd  +\qdd^\T\J_{\phi_i}^\T\A_i\J_{\phi_i}\qdd
\label{eq:appquadraticsumcost3} \\
    &= \frac{1}{2}\sum_i 
(\xdd_i^d)^\T\A_i(\xdd_i^d) 
- 2\qdd^\T\J_{\phi_i}^\T\A_i\xdd_i^d
  +\qdd^\T\J_{\phi_i}^\T\A_i\J_{\phi_i}\qdd,
\label{eq:appquadraticsumcost4}
\end{align}
where the last equality follows from the fact that $a^\T=a$ for any scalar $a$, and $\A_i^\T=\A_i$.  

\subsubsection{Solving the motion generation problem}

First, let us solve the problem using the straightforward approach.  To minimize \myeqref{eq:appquadraticsumcost4}, we simply set the derivative to zero:
\begin{align}
    0 = \frac{\partial c({\qdd})}{\partial \qdd} &= \frac{\partial}{\partial \qdd}
\left(\frac{1}{2}\sum_i 
(\xdd_i^d)^\T\A_i\xdd_i^d 
- 2\qdd^\T\J_{\phi_i}^\T\A_i(\xdd_i^d)  +\qdd^\T\J_{\phi_i}^\T\A_i\J_{\phi_i}\qdd\right) \label{eq:appquadraticsumcostderiv1} \\
 &= \frac{1}{2} \sum_i \frac{\partial}{\partial \qdd}
\big( 
(\xdd_i^d)^\T\A_i(\xdd_i^d) 
- 2\qdd^\T\J_{\phi_i}^\T\A_i\xdd_i^d
+\qdd^\T\J_{\phi_i}^\T\A_i\J_{\phi_i}\qdd\big) \label{eq:appquadraticsumcostderiv2} \\
 &= \frac{1}{2}\sum_i 
- 2\J_{\phi_i}^\T\A_i\xdd_i^d
  +2\J_{\phi_i}^\T\A_i\J_{\phi_i}\qdd \label{eq:appquadraticsumcostderiv3} \\
 &= \sum_i 
  - \J_{\phi_i}^\T\A_i\xdd_i^d  +\J_{\phi_i}^\T\A_i\J_{\phi_i}\qdd.
 \label{eq:appquadraticsumcostderiv4}
\end{align}
Moving terms to both sides of the equation yields
\begin{align}
\sum_i \J_{\phi_i}^\T\A_i\J_{\phi_i}\qdd &=
\sum_i \J_{\phi_i}^\T\A_i\xdd_i^d
 \label{eq:appquadraticsumcostderiv5} \\
\end{align}
which can be solved for the unknown $\qdd$:
\begin{align}
\qdd &= \left(\sum_i  \J_{\phi_i}^\T\A_i \J_{\phi_i}\right)^+\left(\sum_i \J_{\phi_i}^\T\A_i\xdd_i^d\right),  \label{eq:appquadraticsumcostderiv6}
\end{align}
where $^+$ is the pseudoinverse.
The Hessian of 
\myeqref{eq:appquadraticsumcost4} is determined by computing the second derivative:
\begin{align}
    \frac{\partial^2 c({\qdd})}{\partial \qdd^2} &= \frac{\partial^2}{\partial \qdd^2}
\left(\frac{1}{2}\sum_i 
(\xdd_i^d)^\T\A_i\xdd_i^d 
- 2\qdd^\T\J_{\phi_i}^\T\A_i(\xdd_i^d)  +\qdd^\T\J_{\phi_i}^\T\A_i\J_{\phi_i}\qdd\right) \label{eq:appquadraticsumcostderivv1} \\
&= \frac{\partial}{\partial \qdd}
\left(\sum_i 
  - \J_{\phi_i}^\T\A_i\xdd_i^d  +\J_{\phi_i}^\T\A_i\J_{\phi_i}\qdd\right) \label{eq:appquadraticsumcostderivv2} \\
 &= \sum_i 
\J_{\phi_i}^\T\A_i\J_{\phi_i}. \label{eq:appquadraticsumcostderivv3}
\end{align}
In the case that the $\phi_i$s are identical, so that $\J_{\phi_i}=\J$ for all $i$, and assuming that $\J$ is full row rank, then \myeqref{eq:appquadraticsumcostderiv6} simplifies to
\begin{align}
\qdd &= \left(\sum_i  \J^\T\A_i \J\right)^+\left(\sum_i \J^\T\A_i\xdd_i^d\right),  \label{eq:appquadraticsumcostderiv7} \\
&= \J^+\left(\sum_i  \A_i \right)^+\left(\sum_i \A_i\xdd_i^d\right) \label{eq:appquadraticsumcostderiv8} 
\\
&= \J^+\left(\sum_i  \A_i \right)^+\left(\sum_i \A_i\right)\xdd_i^d. \label{eq:appquadraticsumcostderiv9}
\\
&= \J^+\xdd_i^d. \label{eq:appquadraticsumcostderivA}
\end{align}


\subsubsection{Combining RMPs to solve motion generation}

RMPs provide us with a clean mathematical framework in which to reason about motion generation problems such as that above.  In particular, to solve the problem above using RMPs, we simply follow the following steps:  1) create an RMP $(\f_i,\A_i)$ for each task map, where $\f_i \equiv \xdd_i^d$; 2) pull each RMP  back into the configuration space using the pullback operator; and 3) add the pulled-back RMPs using RMP addition.  It is easy to see that, due to the linearity of the RMP operators, the solution from this procedure will be the same as 
\myeqref{eq:appquadraticsumcostderiv6} and
\myeqref{eq:appquadraticsumcostderivv3}, or more specifically, 
\myeqref{eq:appquadraticsumcostderivA} and
\myeqref{eq:appquadraticsumcostderivv3}.
.

\section{Motion generation}
\subsection{handling joint limits}

Define $\sigma : \uu \mapsto \q$ as a  function that satisfies the joint limits by squashing the unconstrained variable $\uu$ through a sigmoid applied dimension-wise, scaled and shifted appropriately so the range spans the robot's joint limits.  That is, $\q = [\begin{matrix}q_1,...,q_d\end{matrix}]^\T = \sigma(\uu) = \sigma([\begin{matrix}u_1,...,u_d\end{matrix}]^\T)$, where $q_i=\sigma_i(u_i)=\alpha_i\frac{1}{1+e^{-u_i}}+\beta_i$, and where $\alpha_i$ and $\beta_i$ are the scaling and offset for the $i^{\text{th}}$ joint.
And denote the Jacobian of that scaled and shifted sigmoid by $\D_\sigma = \frac{\partial \sigma}{\partial \uu}$. This Jacobian is square, diagonal, and invertible for all finite $\uu$; the $i^\text{th}$ element of the diagonal $d_i$ as is given by $d_i = \sigma_i(1-\sigma_i/\alpha_i)$.

Let $\mathcal{R} = {^\calQ}(\f, \A)$ denote an RMP on the configuration space. To regulate joint limits, we add a simple RMP on $\calU$ of the form ${^\calU}(\f_{\text{joint}},\lambda \I)$, where $\lambda$ is a constant. A simple choice of regulating policy is $\f_{\text{joint}} = \zero$, i.e., larger accelerations in $\uu$ are penalized more heavily. By using $\lambda\I$ as a metric, the strength of this penalty is governed by the nonlinearities of $\sigma$. Since 
\begin{align}
    \udd = \D_\sigma^{-1}\qdd\ \ \ \textrm{and}\ \ \  
    \D_\sigma^{-1} = \mathrm{diag}\left(\frac{1}{\sigma_i (1 -\sigma_i/\alpha_i)}\right),
\end{align}
accelerations closer to joint limits are increasingly inflated in $\calU$, thus causing the policy to avoid them.

To see how this new controller affects the original RMP $\calR$, we can use \myeqref{eqn:pullback} to pull it back from $\calU$ to $\calQ$ through the inverse sigmoid map $\uu = \sigma^{-1}(\q)$: 
\begin{align}
\mathtt{pull}_{\sigma^{-1}}    \left({^\calU}(\f_{\text{joint}}, \lambda\I)\right) = {^\calQ}(\D_\sigma \f_{\text{joint}}, \lambda \D_\sigma^{-2})
\label{eqn:jointlimitpull}
\end{align}
and add the result to the original RMP:
\begin{align}
     {^\calQ}(\f', \A') &= {^\calQ}(\f, \A) + {^\calQ}(\D_\sigma \f_{\text{joint}}, \lambda \D_\sigma^{-2}).
\label{eqn:jointlimitsum}
\end{align}
Using \myeqref{eqn:RMPCombinationTwo}, this yields
\begin{align}
     \f' &= (\A + \lambda \D_\sigma^{-2})^{-1}\left(\A\f + \lambda \D_\sigma^{-2}\D_\sigma \f_{\text{joint}}\right)
     \\
     &= \D_\sigma\D_\sigma^{-1}(\A + \lambda \D_\sigma^{-2})^{-1}\D_\sigma^{-1}\D_\sigma\left(\A\f + \lambda \D_\sigma^{-1} \f_{\text{joint}}\right)
     \\
     \label{eqn:GenericJointModf}
     &= \D_\sigma(\D_\sigma\A\D_\sigma + \lambda \I)^{-1}\left(\D_\sigma\A\f + \lambda \f_{\text{joint}}\right) \\
     \A' &= \A + \lambda \D_\sigma^{-2}. \label{eqn:GenericJointModA}
\end{align}

Alternatively, since RMPs are covariant to reparameterization, we can instead pull the RMP $\calR$ from $\calQ$ back into $\calU$:
\begin{align}
    \mathtt{pull}_\sigma\left({^\calQ}(\f, \A)\right)= {^\calU}\left(\D_\sigma^{-1}\f,\ \D_\sigma^\T\A\D_\sigma\right),
\end{align}
which, when added to ${^\calU}(\f_{\text{joint}},\lambda \I)$, yields ${^\calU}(\f',\A')$ such that
\begin{align}
\f'  &= (\D_\sigma\A\D_\sigma + \lambda \I)^{-1}\left(\D_\sigma\A\f + \lambda \f_{\text{joint}}\right) \label{eqn:GenericJointModfU} \\
     \A' &= \D_\sigma\A\D_\sigma + \lambda \I. \label{eqn:GenericJointModAU}.
\end{align}
Note that \myeqref{eqn:GenericJointModfU} is the same as \myeqref{eqn:GenericJointModf} without the $\D_\sigma$ factor, which arises from the relation between the two spaces: $\qdd=\D_\sigma\udd$. 
This demonstrates the covariance of representation (and the independence of the calculations on ordering). Additionally, due to the linearity of the pullback operator, even when operating in $\calU$, other RMPs defined in $\calQ$ can be added by pulling them back into $\calU$ and combining via the metric weighted average without affecting the solution.  Although in the limit, the resulting behavior is the same as Euler integration steps get small, it is often numerically stabler to take finite-sized step through the unconstrained space of $\calU$ rather than the constrained space of $\calQ$.  


If the original RMP $\calR$ is itself the accumulation of multiple RMPs $\calR_1, \calR_2, \ldots$ that were pulled back from $\calX$, then the original controller is given by \myeqref{eqn:pullback} as
\begin{align}
{^\calQ}(\f, \A) &= \sum_i \left(\mathtt{pull}\left({^\calU}(\f_i, \A_i)\right)\right) \\
  &= \sum_i \left({^\calQ}\left(\left(\J_i^\T\A_i\J_i\right)^{+}\J_i^\T\A_i\f_i,\ \J_i^\T\A_i\J_i\right)\right).
\end{align}
Applying \myeqref{eqn:RMPCombination}, this yields
\begin{align}
\f &= \left(\sum_i\J_i^\T\A_i\J_i\right)^+\left(\sum_i\J_i^\T\A_i\J_i\f_i\right) \\
\A &= \sum_i\J_i^\T\A_i\J_i 
\end{align}
Plugging these back into \myeqref{eqn:GenericJointModf}--\eqref{eqn:GenericJointModA} leads to
\begin{align}
\f' &= 
\D_\sigma\left(\D_\sigma\left(\sum_i\J_i^\T\A_i\J_i\right)\D_\sigma + \lambda \I\right)^{-1} \cdot \nonumber \\
    &\ \ \ \ \ \ \ \
    \left(\D_\sigma \sum_i\J_i^\T\A_i\f_i + \lambda \f_{\text{joint}}\right) \\ \label{eqn:JacobianScaling}
    &= \D_\sigma\left(\left(\sum_i\wt{\J}_i^\T\A_i\wt{\J}_i\right) + \lambda \I\right)^{-1}\left(\sum_i\wt{\J}_i^\T\A_i\f_i + \lambda \f_{\text{joint}}\right),
\\
\A' &= \sum_i\J_i^\T\A_i\J_i + \lambda\D_\sigma^{-2},
\end{align}
where $\wt{\J}_i \equiv \J_i\D_\sigma$.  \todo{We probably want to move these derivations to the appendix.}
In other words, handling joint limits requires the differential equation to be augmented in three ways: 1) Scale each Jacobian as $\wt{\J}_i = \J_i\D_\sigma$, which reduces the weights of columns whose joints are near their limit; 2) Add $\lambda \f_{\text{joint}}$ directly to the differential equation, which pushes away from the joint limits if $\f_{\text{joint}} \neq 0$;  and 3) Scale the final resulting acceleration by $\D_\sigma$ to shrink joint accelerations for joints close to their limits.

This technique is smooth, computationally efficient, and numerically stable. It generalizes common Jacobian-pseudoinverse weighting techniques described in the literature \todo{add references} and works well in practice.

\todo{I need to re-read the following paragraph to understand how it fits in.  $\wt{\D}$ is not defined anywhere.  Note that $\alpha_i$ here has a different meaning.  maybe move to appendix.}  Finally, we note that in practice we operate in $\q$ and directly calculate a directionally aware form of the joint limit Jacobian $\wt{\D}_\sigma(\q, \qd)$ as
\begin{align}
    \wt{d}_i = \sigma_i\Big(\alpha_i d_i + \big(1-\alpha_i\big)1\Big)
    + (1-\sigma_i)\Big(\big(1-\alpha_i\big) d_i + \alpha_i1\Big),
\end{align}
where $\sigma_i$ is the joint limit sigmoid of joint $i$ (0 close to the lower limit and 1 close to the upper limit), $\wt{d}_i$ and $d_i$ are the $i^{\mathrm{th}}$ diagonal entries of $\wt{\D}_\sigma$ and $\D_\sigma$, respectively, and $\alpha_i = \sigma(c\dot{q}_i)$ for $c>0$ is a sigmoidal value that is close to 1 for large positive velocities and close to 0 for large negative velocities.
This Jacobian behaves like the sigmoidal Jacobian when moving toward a nearby joint limit, but behaves like the identity Jacobian when moving away from a joint limit. This Jacobian works extremely well in practice even under $\udd_d = \zero$. It's what we use in the experiments.